%% file: neurips_2026.tex
\documentclass{article}

\PassOptionsToPackage{numbers,sort&compress}{natbib}

\usepackage[preprint]{neurips_2026}

\usepackage[utf8]{inputenc} %
\usepackage[T1]{fontenc}    %
\usepackage[pagebackref]{hyperref}       %
\usepackage{url}            %
\usepackage{booktabs}       %
\usepackage{amsfonts}       %
\usepackage{nicefrac}       %
\usepackage{microtype}      %
\usepackage[dvipsnames,table]{xcolor}         %

\usepackage{amsmath}
\usepackage{amssymb}
\usepackage{multirow}
\RequirePackage{xspace}
\usepackage{cleveref}
\usepackage{graphicx}
\usepackage{pifont}
\RequirePackage[font=footnotesize,subrefformat=parens]{subcaption}
\usepackage{wrapfig}

\title{Self-Supervised Learning of Structured Dynamics from Videos}

\author{%
  Lukas Knobel\\
  Fundamental AI Lab, UTN\\
  {\tt\small lukas.knobel@utn.de}\\
  \And
  Andrew Zisserman\\
  VGG, University of Oxford\\
  {\tt\small az@robots.ox.ac.uk}\\
  \And
  Yuki M. Asano\\
  Fundamental AI Lab, UTN\\
  {\tt\small yuki.asano@utn.de}
}

\makeatletter
\DeclareRobustCommand\onedot{\futurelet\@let@token\@onedot}
\def\@onedot{\ifx\@let@token.\else.\null\fi\xspace}
\def\eg{\emph{e.g}\onedot}

 \def\vs{\emph{vs}\onedot}

\makeatother

\newcommand{\method}{SDM\xspace}
\newcommand{\benchmark}{ProbeMotion\xspace}
\newcommand{\cls}{\texttt{CLS}\xspace}
\newcommand{\subtitlerow}[2]{%
    \rowcolor{gray!15}%
    \multicolumn{#1}{@{}l}{\textbf{#2}}\\
}
\definecolor{matchcolour}{rgb}{0.10, 0.9, 0.10}

\definecolor{methodblue}{RGB}{232,242,255}
\newcommand{\std}[1]{{\scriptsize$\pm$#1}}
\newcommand{\gooddelta}[1]{{\color{ForestGreen}\scriptsize~#1}}
\newcommand{\baddelta}[1]{{\color{red}\scriptsize~#1}}
\colorlet{strongsupervisedgray}{black!55}
\newcommand{\strongsupervisedstyle}[1]{\textcolor{strongsupervisedgray}{#1}}
\newcommand{\strongsupervisedrow}[8]{%
\strongsupervisedstyle{#1} &
\strongsupervisedstyle{#2} &
\strongsupervisedstyle{#3} &
\strongsupervisedstyle{#4} &
\strongsupervisedstyle{#5} &
\strongsupervisedstyle{#6} &&
\strongsupervisedstyle{#7} &
\strongsupervisedstyle{#8}
}
\newcommand{\smallparagraph}[1]{\noindent\textbf{#1.}}

\definecolor{res}{RGB}{248,219,215}
\definecolor{prim}{RGB}{255,244,204}
\usepackage{calc}

\newlength{\DepthReference}
\newlength{\HeightReference}
\newlength{\Width}%

\newcommand{\MyColorBox}[2][red]%
{%
    \settowidth{\Width}{#2}%
    \colorbox{#1}%
    {%
        \raisebox{-\DepthReference}%
        {%
                \parbox[b][\HeightReference+\DepthReference][c]{\Width}{\centering#2}%
        }%
    }%
}

\newcommand{\cmark}{\ding{51}}
\newcommand{\xmark}{\ding{55}}

\begin{document}

\maketitle

\input{sections/0_abstract}    
\input{sections/1_intro}
\input{sections/3_method}
\input{sections/4_experiments}
\input{sections/2_related_work}
\input{sections/5_conclusion}

\begin{ack}
LK is grateful for support through the ELLIS PhD Program.
The authors gratefully acknowledge the scientific support and HPC resources provided by the Erlangen National High Performance Computing Center (NHR@FAU) of the Friedrich-Alexander-Universität Erlangen-Nürnberg (FAU) under the BayernKI project $\mathrm{v115be}$. BayernKI funding is provided by Bavarian state authorities. The research is also supported by the UK EPSRC Programme Grant VisualAI (EP/T028572/1) and a Royal Society Research Professorship RSRP$\backslash$R$\backslash$241003.
\end{ack}

\bibliographystyle{plainnat}
\bibliography{references}

\newpage
\appendix

\input{sections/X_suppl}

\end{document}

%% file: sections/0_abstract.tex
\begin{abstract}
Understanding motion in video is a fundamental challenge for visual learning, as frame-to-frame change entangles two sources of dynamics: camera motion and object motion. This decomposition has remained underexplored in representation learning, partly because these factors are tightly coupled in natural videos and difficult to supervise separately. Yet recovering it is important for learning robust motion representations that separate meaningful object dynamics from camera-induced variation.
We study whether such structured motion representations can be recovered from frozen features of a pretrained image vision transformer. 
We propose the Structured Dynamics Model (SDM), which explicitly separates the dominant source of temporal change from residual dynamics through future-feature prediction, rather than representing video change with a single entangled latent or with unstructured, spatially dense transition tokens.
Training combines self-supervised learning on real video with weak supervision of scene dynamics on synthetic Kubric data.
We evaluate SDM on ProbeMotion, a new evaluation suite spanning synthetic and real videos with camera motion, object motion, and combined dynamics. SDM outperforms backbone baselines using global \texttt{CLS} or average-pooled features, and compares favorably to strongly supervised representations such as VGGT on several probes, despite using {\em substantially} weaker supervision. These results suggest that pretrained image models can be readily repurposed into structured video-dynamics representations, providing a useful inductive bias for learning and analyzing latent video dynamics.\footnote{Project page: \href{https://lukasknobel.github.io/projects/StructuredDynamics}{https://lukasknobel.github.io/projects/StructuredDynamics}}

\end{abstract}

%% file: sections/1_intro.tex
\section{Introduction}
\label{sec:intro}
Recent progress in computer vision has produced strong self-supervised image backbones that capture rich semantic and geometric structure from large-scale image data~\cite{oquabdinov2,simeoni2025dinov3,weinzaepfel2022croco,tschannen2025siglip2,cao2026tipsv2,yang2025pixio,fini2025aimv2}. 
These representations are increasingly used as foundations for 3D perception and reconstruction systems. However, despite building on self-supervised backbones, many successful approaches still rely on substantial supervision during adaptation, for example through 3D annotations, pseudo-ground truth from structure-from-motion pipelines, or other geometric labels~\cite{wang2025vggt,wang2024dust3r,wang2025pi3,lin2025depthanythingv3,luo20264rc}. VGGT, for instance, trains on 17 annotated datasets~\cite{wang2025vggt}. This limits scalability and leaves open the question of how much scene dynamics can be recovered directly from generic pretrained visual features without relying on dense supervision.

Video offers a natural source of supervision for this problem. However, frame-to-frame change entangles the two sources of dynamics, camera motion and object motion.
Standard self-supervised prediction objectives can therefore model aggregate change without separating its underlying causes.
Recovering such structure is important for learning motion representations organized around interpretable physical factors, but remains underexplored in representation learning. In particular, many latent action or world models summarize scene change with a single latent token~\cite{garrido2026latentinthewild,kerssies2026frame}, mixing dominant global change with residual object-centric dynamics.

Rather than training a video model end-to-end, we ask whether a pretrained image backbone already contains enough information to support an explicit structuring of scene dynamics. 
To this end, we build a {\em structured dynamics} model (\method) on top of frozen backbone features and train it through future feature prediction. 
We find that, when combined with weak supervision on synthetic data indicating only whether the camera or the scene is static, \method organizes temporal change into primary and residual motion components that adapt to the dominant sources of change in each video.

We evaluate this setting on an evaluation suite of benchmarks spanning synthetic and real videos with camera motion, object motion, and combined dynamics. We refer to this collective evaluation as \benchmark. Across these settings, \method consistently outperforms na\"ive baselines based on global \cls or average-pooled features and even surpasses VGGT-based probes on $3$ out of $7$ benchmarks.
These results suggest that pretrained image representations can be remodeled into strong structured video-dynamics representations, with only weak synthetic supervision and self-supervised learning on unlabeled real videos.

Our contributions are:
\begin{itemize}
\itemsep0.5em
    \item We study whether structured motion representations can be extracted from frozen pretrained image features.
    \item We introduce the Structured Dynamics Model (\method), a simple architecture on top of self-supervised image backbones that learns this structure through future feature prediction, outperforms na\"ive global-feature baselines, and is competitive with strongly supervised 3D representations on several motion-probing tasks.
    \item We develop \benchmark, an evaluation suite spanning synthetic and real videos covering camera motion, object motion, and combined dynamics.
\end{itemize}

\begin{figure}[t]
  \centering
  \includegraphics[width=\textwidth]{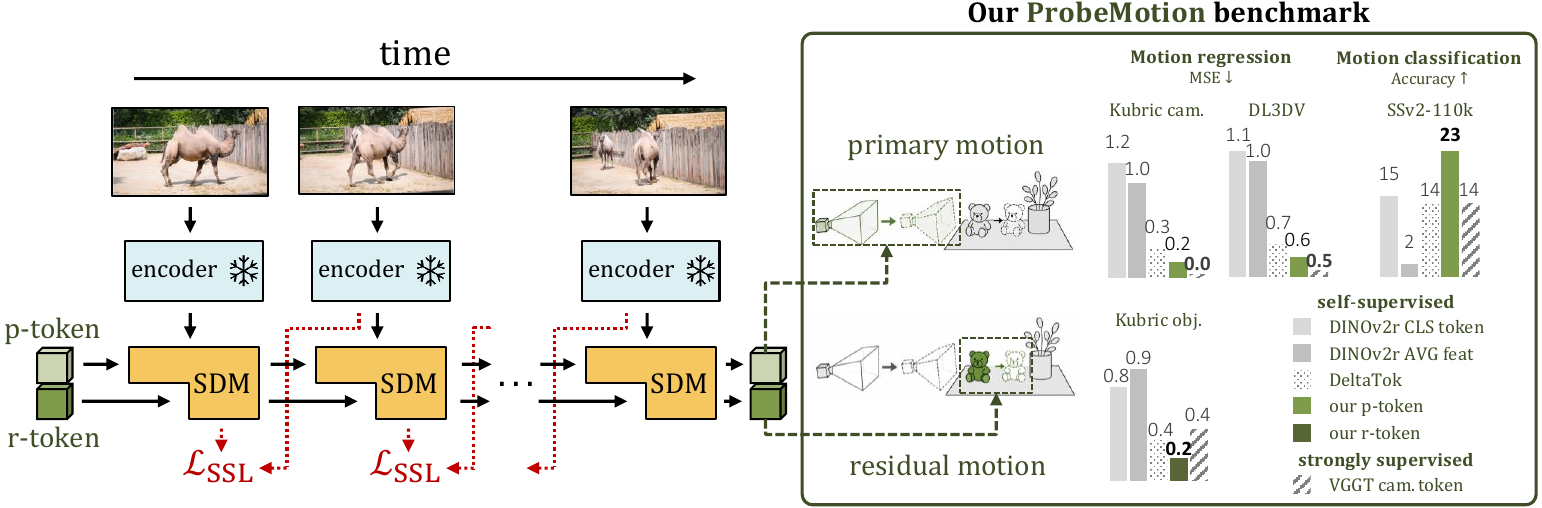}
  \caption{\textbf{The Structured Dynamics Model (\method) remodels frozen image features into structured motion tokens.}
    Left: A frozen image encoder extracts frame features, and a recurrent \method is trained by self-supervised future-feature prediction.
    \method structures temporal change into a primary motion token $p$ for the dominant source of change and a residual motion token $r$ for remaining dynamics.
    Right: Linear probes on \benchmark show that these structured motion tokens outperform direct frozen-backbone descriptors such as \cls and average-pooled features, 
    and even surpass VGGT probing performance on 3/7 tasks (see~\Cref{tab:baselines}), while showing stronger generalization to motion-adjacent semantic action classification.
  }
  \label{fig:overview}
  \vspace{-1em}
\end{figure}

%% file: sections/3_method.tex
\section{Method}
\label{sec:method}

We learn structured motion representations by predicting how frozen visual features change over time. Rather than training a video encoder, we ask whether a simple model, \method, on top of a pretrained image backbone can organize temporal change into primary and residual motion components, which explain the dominant source of scene change and remaining dynamics. An overview is shown in \Cref{fig:method}.

\subsection{Frozen visual features}

Let $\mathcal{E}$ denote a frozen vision transformer. For each frame $x_t$, we extract a spatial feature map:
\begin{equation}
    f_t = \mathcal{E}(x_t) \in \mathbb{R}^{H \times W \times D},
\end{equation}
where $H,W$ are spatial dimensions and $D$ is the feature dimension. \method predicts the target feature map $f_t$ from $f_{t-1}$ in frozen feature space. The model is autoregressive in its recurrent motion tokens but non-causal, $f_t$ is available for extracting the transition token used to predict $f_t$ from $f_{t-1}$.

Conditional feature prediction requires information about the target transition. CroCo and SiamMAE use cross-view conditioning~\cite{weinzaepfel2022croco,gupta2023siammae}, while latent-action models use compact transition tokens for future prediction~\cite{garrido2026latentinthewild,kerssies2026frame}. A single transition token must explain all sources of change, from appearance-dependent feature variation to low-dimensional motion factors such as camera translation or object direction. We instead follow the intuition that much of the temporal change in videos is governed by lower-dimensional dynamics acting on high-dimensional visual features. This motivates a \emph{structured} prediction model that organizes temporal change into primary and residual motion tokens.

\subsection{Structured dynamics model}\label{sec:structured_prediction}
\begin{figure}[ht]
  \centering
  \includegraphics[width=\textwidth]{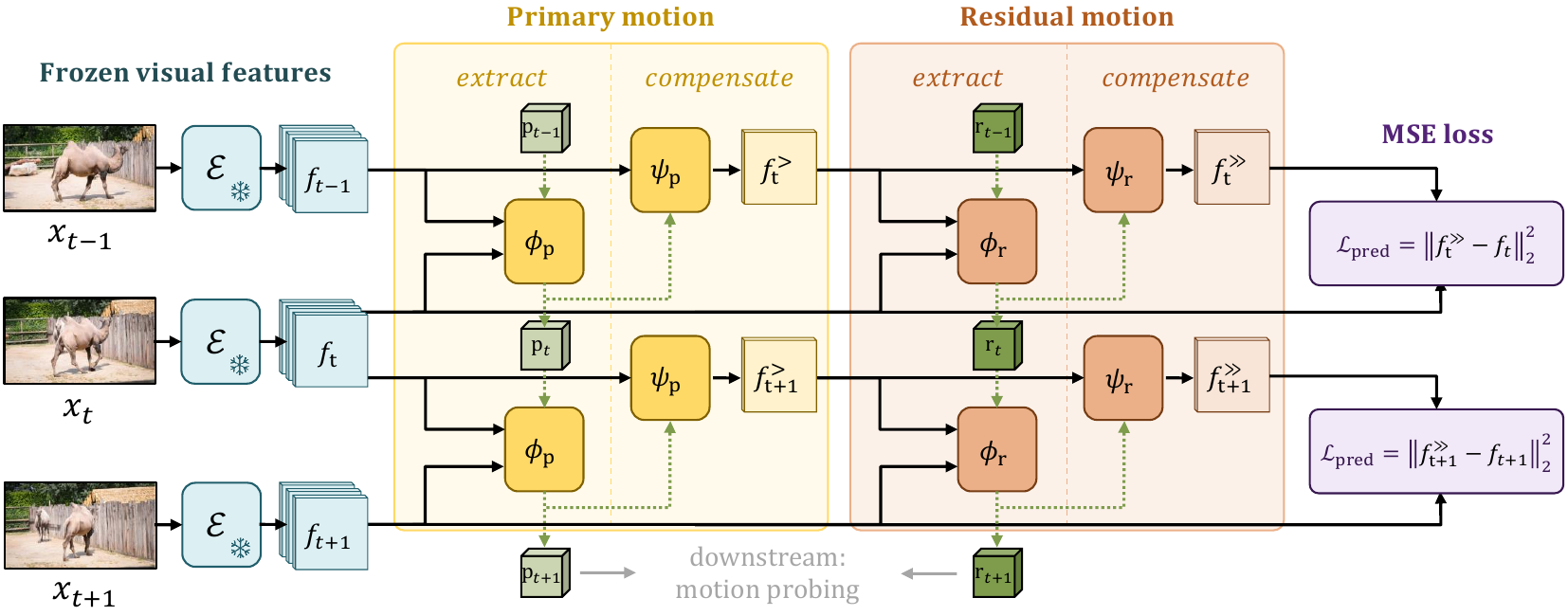}
  \caption{\textbf{Overview of the structured dynamics model shown for T=3.}
For each transition, SDM predicts the target frozen feature map $f_t$ from the previous one $f_{t-1}$ through two sequential compensation stages. The \emph{primary} stage extracts a recurrent motion token $p_t$ from the feature pair $(f_{t-1},f_t)$ and uses it to produce a primary-compensated feature map $f^{>}_t$, explaining the dominant change. The \emph{residual} stage extracts $r_t$ from the mismatch between $f^{>}_t$ and $f_t$, and refines $f^{>}_t$ into the final prediction $f^{\gg}_t$. The motion tokens are autoregressive but non-causal across time, allowing temporal information to accumulate.
}
  \label{fig:method}
\end{figure}

Figure~\ref{fig:method} gives an overview of \method. Given consecutive feature maps $(f_{t-1},f_t)$, \method predicts $f_t$ through two sequential stages. The \emph{primary stage} consists of a motion extractor $\phi_p$ and a feature predictor $\psi_p$. The extractor reads the current feature pair $(f_{t-1}, f_t)$ and updates the previous primary token $p_{t-1}$, giving $p_t$. The predictor then uses $p_t$ to compensate the source feature map $f_{t-1}$, producing an intermediate prediction $f^{>}_t$, pronounced ‘f modified once’. The \emph{residual stage} has the same structure, with modules $\phi_r$ and $\psi_r$. It estimates a residual token $r_t$ from the remaining mismatch between $f^{>}_t$ and $f_t$, and uses this token to refine $f^{>}_t$ into the final prediction $f^{\gg}_t$, pronounced ‘f modified twice’.

\smallparagraph{Primary motion extraction}
The primary extractor updates a recurrent motion token $p_t \in \mathbb{R}^D$ by attending to the current frame pair:
\begin{equation}
    p_t = \phi_p(p_{t-1}; f_{t-1}, f_t),
\end{equation}

where $p_0$ is learnable.
The recurrence allows the model to integrate motion information over time while each update only compares adjacent frames.
The resulting token is then used to compensate the source feature map.

\smallparagraph{Primary motion compensation}
The predictor $\psi_p$ uses $p_t$ to compensate the source feature map:
\begin{equation}
    f^{>}_t = \psi_p(f_{t-1}; p_t).
\end{equation}

We interpret $f^{>}_t$ as a primary-motion-compensated intermediate representation that serves as input to the residual stage. 
It is trained to explain the part of the transition from $f_{t-1}$ to $f_t$ that can be captured by the primary source of scene change.

\smallparagraph{Residual motion extraction}
Given $f^{>}_t$, the remaining discrepancy to $f_t$ corresponds to dynamics not explained by the primary component.
We capture this with a second latent token $r_t \in \mathbb{R}^D$:
\begin{equation}
    r_t = \phi_r(r_{t-1}; f^{>}_t, f_t),
\end{equation}

where $r_0$ is learnable. Conditioning it on $f^{>}_t$ encourages $r_t$ to represent only the residual dynamics that remain after the primary component has been accounted for.

\smallparagraph{Residual motion compensation}
The residual predictor refines the intermediate representation:
\begin{equation}
    f^{\gg}_t = \psi_r(f^{>}_t; r_t).
\end{equation}

The final prediction $f^{\gg}_t$ is therefore obtained by first compensating for primary and then residual dynamics.
The pair $(p_t, r_t)$ forms the structured motion representation used throughout the paper.

\smallparagraph{Extension beyond pairs of frames}
\method naturally extends to longer clips. We apply the same updates at each timestep and carry $p_{t-1}$ and $r_{t-1}$ forward. Thus, temporal context is accumulated in the recurrent motion tokens.

\subsection{Training objective}\label{sec:training_objective}
\method is trained through feature prediction in the frozen encoder space.
To align the architectural decomposition with the intended semantics of primary and residual motion, we use two weak scene-level annotations, static scene and static camera, and selectively apply the loss.

\smallparagraph{Unlabeled or dynamic videos}
For unlabeled videos and samples with both camera and scene dynamics, we supervise only the final prediction: $\mathcal{L}
=\mathcal{L}_{\mathrm{MSE}}(f^{\gg}_t, f_t)$, where $\mathcal{L}_{\mathrm{MSE}}(x,y)=\frac{1}{HWD} \left\| x - y \right\|_2^2$. 

\smallparagraph{Static scenes}
For static-scene samples, the primary component should explain the full transition, so we bypass the residual stage and supervise the output of the primary compensation directly: $\mathcal{L}
=\mathcal{L}_{\mathrm{MSE}}(f^{>}_t, f_t)$.

\smallparagraph{Static cameras}
If a sample is annotated as having a static camera, there is no camera-induced global change to compensate for.
Hence, the primary stage is regularized to leave the source representation unchanged while the residual stage captures scene dynamics: $\mathcal{L}
=\mathcal{L}_{\mathrm{MSE}}(f^{\gg}_t, f_t)+\lambda_{\mathrm{reg}}\mathcal{L}_{\mathrm{MSE}}(f^{>}_t, f_{t-1})$.

\subsection{Implementation details}\label{sec:implementation_details}

\smallparagraph{Visual features}
We use frozen DINOv2-B/14 with registers~\cite{darcet2024dino_registers} as $\mathcal{E}$ and extract patch features from all $12$ blocks. 
Following \citet{mur2026vjepa21}, we concatenate these intermediate features channel-wise and project them with a two-layer MLP to $D{=}768$, matching the hidden dimension of one backbone block, before applying \method.
The model's predictions are then linearly projected back to the concatenated multi-layer feature space before applying the loss.

\smallparagraph{Motion extractor $\phi$}
Both $\phi_p$ and $\phi_r$ are $4$-block transformer decoders with $n_{\text{reg}}=4$ learnable registers~\cite{darcet2024dino_registers}. 
The motion token and registers self-attend and cross-attend to frame features, using a 3D extension of RoPE~\cite{su2024rope,assran2025vjepa2} for spatial and temporal positions with a base of $100$~\cite{heo2024rope_vit}. 
The initial primary and residual motion tokens are learnable parameters initialized from $\mathcal{N}(0,\ 0.02^2)$.

\smallparagraph{Predictor $\psi$}
We parameterize $\psi_p$ and $\psi_r$ as shallow $2$-block decoders that cross-attend to the corresponding motion tokens. We use 2D RoPE for spatial positions~\cite{heo2024rope_vit}.

%% file: sections/4_experiments.tex
\section{Experiments}
\label{sec:experiments}
\subsection{Datasets}\label{sec:datasets}

We train on a mixture of synthetic and real videos: Kubric~\cite{greff2021kubric}, SSv2~\cite{goyal2017ssv2}, and DL3DV~\cite{ling2024dl3dv}. Kubric provides weak scene-level labels indicating whether the camera or scene is static, while SSv2 and DL3DV are used without labels. 
We generate $180$k Kubric sequences, equally split between static-scene and dynamic-scene videos. Static-scene videos use a moving camera, while dynamic-scene videos are further equally split between moving-camera and static-camera settings (see \Cref{sec:kubric_details_appendix} for details).
Real-video training uses approximately $170$k SSv2 clips and $4$k DL3DV clips.

For evaluation, we introduce \benchmark, a diverse benchmark suite for probing structured motion representations across Kubric, DL3DV, CameraBench~\cite{lin2025camerabench}, static-camera subsets of DAVIS2017~\cite{pont2017davis2017} and YouTubeVOS~\cite{xu2018youtubevos1,xu2018youtubevos2}, and an SSv2 subset denoted SSv2-$110$k. \Cref{tab:datasets} summarizes the datasets. Split and sampling details are in \Cref{sec:data_split_sampling_appendix}.

Kubric evaluates controlled synthetic camera and object motion with disjoint training/test objects and backgrounds as well as randomly sampled motions. DL3DV and CameraBench evaluate real camera motion, while static DAVIS2017 and static YouTubeVOS evaluate 2D object motion under approximately static cameras. We construct the static-camera subsets by filtering clips using VGGT-estimated camera motion and labeling 2D object motion from mask-centroid displacement (see \Cref{sec:static_davis_youtubevos_construction_appendix} for details). SSv2-$110$k evaluates motion-heavy action semantics.

\begin{table*}[t]
    \caption{\textbf{\benchmark evaluation datasets.} Summary of the linear-probe datasets. Numbers given for source videos refer to the samples taken from the corresponding public or generated dataset after clip-length filtering. Dashes indicate that the source dataset does not provide a split suitable for \benchmark. For these datasets, we create our own probing splits. Parentheses indicate partial dynamics for SSv2's camera (see \Cref{sec:ssv2_camera_motion_appendix}). Further details, including split creation for these datasets, are provided in \Cref{sec:data_split_sampling_appendix}.}
    \label{tab:datasets}
    \centering
    \small
    \setlength{\tabcolsep}{0.4em}
    \begin{tabular}{l | cccccc }
        \toprule
        & \multirow{2}*{Kubric} & \multirow{2}*{DL3DV} & Static & Static & Camera & SSv2-\\
        & & & DAVIS2017 & YouTubeVOS & Bench & $110$k\\
        \midrule
        Source train videos & $15$k & $3.5$k & $60$ & $3.5$k  & -- & $110$k\\
        Source test videos & $5$k & -- & $30$ & --  & $1$k & $20$k\\
        Moving camera & \cmark & \cmark & \xmark & \xmark & \cmark & (\xmark) \\
        Dynamic scene & \cmark & \xmark & \cmark & \cmark & \cmark & \cmark \\
        Num. frames per clip & 4 & 5 & 4 & 4 & 5 & 7\\
        \multirow{2}*{Probing target} & cam.+obj. & cam. & obj. & obj. & cam. motion & action\\
         & motion & motion & displacement & displacement & class & class\\
        \bottomrule
    \end{tabular}
\end{table*}

\subsection{Training and evaluation details}\label{sec:training_eval_details}
\smallparagraph{Training details}
Mini-batches mix Kubric, SSv2, and DL3DV with fixed ratios of $0.5/0.25/0.25$. 
Kubric samples are drawn from the three controlled settings described above: moving-camera/static-scene, static-camera/dynamic-scene, and moving-camera/dynamic-scene. Consequently, $1/4$ of each mini-batch receives static-scene supervision, $1/8$ receives static-camera supervision, and the remaining $5/8$ is trained only with the standard final-prediction objective.
This last group consists of moving-camera/dynamic-scene Kubric samples and all unlabeled real-video samples, for which we do not impose additional structure on the decomposition. 
We sample $5$ frames at $2$ fps and resize inputs to $224{\times}224$ without additional data augmentation.
We train all components with AdamW for $200$k iterations, batch size $128$, learning rate $6.25{\times}10^{-5}$ with $1$k warmup, weight decay $0.05$, and set the regularization weight to $\lambda_{\mathrm{reg}}{=}0.5$. One run takes approximately $11$h on $4$ H100 GPUs.

\smallparagraph{Evaluation protocol}
We evaluate \benchmark with linear probes on frozen \method motion tokens. 
Unless noted otherwise, we probe $p$ for camera motion in Kubric, DL3DV, and CameraBench, and for object motion in static DAVIS2017/YouTubeVOS, where object motion is the dominant source of change. 
A VGGT-based analysis indicates low inter-frame camera motion in SSv2-$110$k (see \Cref{sec:ssv2_camera_motion_appendix}), so we follow the same reasoning and probe $p$ for action recognition.
For Kubric object motion in dynamic-camera scenes, we probe $r$, and study token swapping in \Cref{sec:analysis_of_sdm}. 
We report final-timestep performance and standard deviations over $3$ seeds.
We use linear classifiers for discrete labels and linear regressors for continuous motion targets.
Kubric and DL3DV use normalized-MSE regression on 6-DoF camera motion, with Kubric object motion evaluated as 3D translation, since the simulated objects do not undergo rotation. Static DAVIS2017 and YouTubeVOS use 2D object-displacement regression from mask-centroid motion. CameraBench and SSv2-$110$k use top-1 classification accuracy.
Full task definitions are in \Cref{tab:evaluation_tasks} in the appendix.
All probes are single linear layers trained on frozen features. Classification sets are class-balanced by downsampling, and regression sets use motion-magnitude subsampling to reduce the number of lower-motion samples. 
All methods use the same protocol and report the best held-out metric over probe training to reduce sensitivity to probe optimization hyperparameters (see \Cref{sec:data_eval_details_appendix}).

\subsection{Comparison to baselines} \label{sec:comparison_baselines}

\smallparagraph{Baselines}
Since image backbones produce frame-level representations, we convert consecutive-frame features into motion descriptors using either concatenation or feature differences.
We compare against direct frozen-backbone motion descriptors derived from either the \cls{} token or average-pooled patch features. Based on descriptor ablations (see \Cref{sec:baseline_results_appendix}), we use concatenation for \cls{} features and feature differences for AVG-pooled features.
Multi-frame variants average these frame-pair descriptors over time to match \method's temporal context. 
We additionally evaluate the self-supervised DeltaTok~\cite{kerssies2026frame}, which represents frame-to-frame change with a single delta token. Compared with \method, DeltaTok is trained for $8{\times}$ more iterations and operates on higher-resolution inputs, making it a strong baseline (see \Cref{sec:deltatok_eval_appendix}).
We also compare to strongly supervised 3D and geometry models: VGGT-$1$B, Depth Anything 3 (DA3)~\cite{lin2025depthanythingv3}, and Pi3X~\cite{wang2025pi3}. 
For each model, we report separate rows for descriptors derived from average-pooled feature maps and camera tokens, selecting the better of concatenation and feature differences for each benchmark based on \Cref{sec:baseline_results_appendix}. Since Pi3X does not provide camera tokens, we use averaged register tokens as the corresponding descriptor.
These models are substantially larger than \method (roughly $1$B parameters each \vs $215$M parameters, including an $87$M frozen backbone) and are trained with explicit geometric supervision, such as camera pose, depth, or point clouds, and are therefore included to contextualize performance under much stronger supervision.
\begin{table*}[t]
    \caption{\textbf{Linear probing results on \benchmark.}
        We compare \method against direct frozen-backbone descriptors from DINOv2 with registers, the self-supervised DeltaTok model~\cite{kerssies2026frame}, and strongly supervised 3D/geometry models.
        For the direct DINOv2 baselines, we use concatenation for \cls{} features and feature differences for AVG-pooled features. 
        For supervised models, we report separate rows for average-pooled feature-map descriptors and camera-token descriptors. Each entry selects the better of concatenation and feature differences for that row and benchmark based on \Cref{sec:baseline_results_appendix}. For Pi3X, which does not provide camera tokens, the second row uses averaged register tokens.
        For DeltaTok, we directly probe its delta token.
        \method outperforms the direct frozen-backbone descriptors across most camera-motion, object-motion, and action probes, and outperforms DeltaTok on five of seven tasks, while remaining competitive with strongly supervised representations despite using substantially weaker supervision and fewer parameters.
        We report MSE on standardized regression targets and accuracy in \% for classification tasks.
        Best and second-best results among methods without explicit geometric supervision are highlighted in \textbf{bold} and \underline{underlined}, respectively.
    }
    \label{tab:baselines}
    \centering
    \small
    \setlength{\tabcolsep}{0.35em}
    \begin{tabular}{l ccc c cccc }
        \toprule
        \multirow{3}*{Method} & \multicolumn{5}{c}{Regression (MSE $\downarrow$)} && \multicolumn{2}{c}{Classification (accuracy $\uparrow$)}\\
        \cmidrule{2-6} \cmidrule{8-9}
        &\multicolumn{2}{c}{Kubric} & \multirow{2}*{DL3DV} & \multicolumn{1}{c}{Static} & \multicolumn{1}{c}{Static} && \multicolumn{1}{c}{Camera} & \multicolumn{1}{c}{SSv2-$110$k}\\
        & \multicolumn{1}{c}{cam} & \multicolumn{1}{c}{obj}& & \multicolumn{1}{c}{DAVIS2017} & \multicolumn{1}{c}{YouTubeVOS} && \multicolumn{1}{c}{Bench} & (action pred.) \\
        \midrule
        \subtitlerow{9}{strongly supervised}
        \strongsupervisedrow{VGGT avg. feat.} {0.09\std{0.0}} {0.32\std{0.0}} {0.56\std{0.0}} {0.78\std{0.1}} {0.79\std{0.0}} {82.9\std{0.0}} {15.5\std{0.1}}\\
        \strongsupervisedrow{VGGT cam. token} {0.03\std{0.0}} {0.43\std{0.0}} {0.53\std{0.0}} {0.71\std{0.0}} {0.81\std{0.0}} {88.7\std{0.7}} {13.5\std{0.1}}\\
        \strongsupervisedrow{DA3 avg. feat.} {0.11\std{0.0}} {0.40\std{0.0}} {0.55\std{0.0}} {0.87\std{0.1}} {0.89\std{0.0}} {83.3\std{0.4}} {14.0\std{0.1}}\\
        \strongsupervisedrow{DA3 cam. token} {0.04\std{0.0}} {0.58\std{0.0}} {0.53\std{0.0}} {0.87\std{0.0}} {0.91\std{0.0}} {84.4\std{1.3}} {12.2\std{0.4}}\\
        \strongsupervisedrow{Pi3X avg. feat.} {0.07\std{0.0}} {0.38\std{0.0}} {0.52\std{0.0}} {0.52\std{0.0}} {0.75\std{0.0}} {87.9\std{0.7}} {19.8\std{0.1}}\\
        \strongsupervisedrow{Pi3X avg. regs.} {0.08\std{0.0}} {0.45\std{0.0}} {0.53\std{0.0}} {0.61\std{0.0}} {0.77\std{0.0}} {85.7\std{0.4}} {12.7\std{0.3}}\\
        \midrule
        \subtitlerow{9}{baselines}
        DeltaTok & \underline{0.29}\std{0.0} & \underline{0.35}\std{0.0} & \underline{0.65}\std{0.0} & \textbf{0.69}\std{0.0} & \underline{0.75}\std{0.0} && \textbf{89.6}\std{0.9} & 13.5\std{0.2}\\
        AVG-pool & 0.96\std{0.0} & 0.93\std{0.0} & 1.04\std{0.0} & \underline{0.78}\std{0.0} & 0.91\std{0.0} && 67.0\std{1.2} & 2.3\std{0.0}\\
        \cls & 1.16\std{0.0} & 0.78\std{0.0} & 1.09\std{0.0} & 0.90\std{0.1} & 1.05\std{0.0} && 68.6\std{0.8} & \underline{14.9}\std{0.2}\\
        \midrule
        \rowcolor{methodblue}
        \method & \textbf{0.16}\std{0.0} & \textbf{0.19}\std{0.0} & \textbf{0.59}\std{0.0} & 0.87\std{0.1} & \textbf{0.71}\std{0.0} && \underline{85.3}\std{1.7} & \textbf{23.1}\std{1.3}\\
        \bottomrule
    \end{tabular}
\end{table*}

\smallparagraph{\method outperforms direct frozen-backbone descriptors}
Across \benchmark, \method improves over \cls{} and AVG-pool baselines on all tasks except static DAVIS2017, where AVG-pool performs best on 2D object-displacement regression.
This exception might reflect the small, curated nature of static DAVIS2017 and the relative simplicity of image-plane displacement prediction. The direct baselines remain substantially weaker on 3D motion probes such as Kubric and DL3DV, suggesting that gains come from structured future-feature prediction rather than strong frozen features alone. This remains true when the direct baselines are given access to multi-layer DINOv2 features from all $12$ blocks (see \Cref{tab:multi_layer_baseline_ablation_appendix}). 
The improvement also transfers beyond datasets used for unlabeled \method training, \eg static YouTubeVOS improves over AVG-pool by reducing MSE by $0.2$.

\smallparagraph{\method outperforms DeltaTok across most probes despite a smaller training budget}
\method outperforms DeltaTok on five of seven \benchmark tasks, with the largest gains on Kubric object motion ($0.16$ lower MSE) and SSv2-$110$k ($9.6$ p.p.\ higher accuracy). DeltaTok performs better on static DAVIS2017 and CameraBench. Overall, these results suggest that structured latents provide strong performance across a broad range of motion and action probes.

\smallparagraph{\method is competitive with supervised 3D representations on \benchmark}
Despite using weaker supervision and fewer parameters than the supervised 3D and geometry baselines, \method approaches supervised-model performance on DL3DV probing while performing similarly on CameraBench.
Furthermore, \method performs best on two out of three object-motion probes, improving over the strongest supervised baseline by $0.13$ MSE on Kubric object motion and $0.04$ MSE on static YouTubeVOS. On static DAVIS2017, \method is on par with DA3, although VGGT and Pi3X perform better.
This suggests that structured feature prediction recovers motion-relevant information not fully captured by supervised descriptors alone.

\smallparagraph{Primary tokens generalize to object-centric action prediction}
On SSv2-$110$k, \method improves over \cls{} and AVG-pool by $8.2$ and $20.8$ p.p., respectively, exceeds DeltaTok by $9.6$ p.p., and remains above the best supervised descriptor, Pi3X average-feature concatenation, by $3.3$ p.p. even though the feature concatenation doubles the linear probe's input dimension compared to \method's probe.

\subsection{Analysis of SDM}\label{sec:analysis_of_sdm}
\begin{table*}[t]
\centering
\begin{minipage}[t]{0.60\textwidth}
    \centering
    \caption{\textbf{SDM utilizes temporal context to aggregate motion information.}
        We compare a single frame pair ($T=2$) against the maximum context for Kubric and CameraBench, and evaluate on sequences longer than the training range of $T=5$ on SSv2-$110$k. The baseline is the AVG-pooled feature-difference descriptor averaged over frame pairs. We report MSE on standardized frame-to-frame pose deltas for regression on Kubric and accuracy in \% for CameraBench and SSv2-$110$k.
    }
    \label{tab:temporal_context}
    \small
    \setlength{\tabcolsep}{0.35em}
    \begin{tabular}{l| cc cc cc}
        \toprule
        \multirow{2}*{method} 
        & \multicolumn{2}{c|}{Kubric cam. $\downarrow$}
        & \multicolumn{2}{c|}{Kubric obj. $\downarrow$}
        & \multicolumn{2}{c}{CameraBench $\uparrow$}\\
        & T=2 & \multicolumn{1}{c|}{T=4} & T=2 & \multicolumn{1}{c|}{T=4} & T=2 & T=5\\
        \midrule
        baseline & 0.96 & \multicolumn{1}{c|}{0.87\gooddelta{-0.09}} & 0.93 & \multicolumn{1}{c|}{0.72\gooddelta{-0.21}} & 67.1 & 69.2\gooddelta{+2.1}\\
        \rowcolor{methodblue}\method & 0.26 & \multicolumn{1}{c|}{0.16\gooddelta{-0.10}} & 0.22 & \multicolumn{1}{c|}{0.19\gooddelta{-0.03}} & 84.7 & 85.3\gooddelta{+0.6}\\
        \midrule\midrule
        \multirow{2}*{method} 
        & \multicolumn{6}{c}{SSv2-$110$k $\uparrow$}\\
        & T=2 & T=3 & T=4 & T=5 & T=6 & T=7\\
        \midrule
        baseline & 2.4 & 5.9\gooddelta{+3.5} & 8.4\gooddelta{+2.5} & 10.9\gooddelta{+2.5} & 11.7\gooddelta{+0.8} & 13.3\gooddelta{+1.6}\\
        \rowcolor{methodblue}\method & 16.7 & 18.6\gooddelta{+1.9} & 20.4\gooddelta{+1.8} & 21.6\gooddelta{+1.2} & 22.5\gooddelta{+0.9} & 23.1\gooddelta{+0.6}\\
        \bottomrule
    \end{tabular}
\end{minipage}
\hfill
\begin{minipage}[t]{0.37\textwidth}
    \centering
    \caption{\textbf{SDM works across backbones.}
        We report DL3DV MSE on standardized frame-to-frame pose deltas and CameraBench / SSv2-$110$k accuracy in \%. MAE~\cite{he2022mae}, SigLIP~\cite{zhai2023siglip}, and DINOv3~\cite{simeoni2025dinov3} use the base variant with patch size $16$. DINOv2 with registers~\cite{darcet2024dino_registers} (DINOv2r) uses patch size $14$. 
    }
    \label{tab:backbone_ablation}
    \small
    \setlength{\tabcolsep}{0.35em}
    \begin{tabular}{l | ccc}
        \toprule
        \multirow{2}*{backbone} 
        & \multirow{2}*{DL3DV$\downarrow$} 
        & Camera & SSv2\\
        & & Bench $\uparrow$ & $110$k$\uparrow$\\
        \midrule
        MAE & 0.61 & 87.6 & 14.1\\
        SigLIP & 0.64 & 85.1 & 11.6\\
        DINOv3 & 0.58 & 86.8 & 27.0\\
        \rowcolor{methodblue}DINOv2r & 0.59 & 85.3 & 23.1\\
        \bottomrule
    \end{tabular}
\end{minipage}
\end{table*}
\smallparagraph{Utilizing temporal context}
\Cref{tab:temporal_context} evaluates whether \method can exploit longer temporal context when predicting the final timestep.
\method benefits from additional context on temporally coherent benchmarks. 
On Kubric, where camera and object motion are linear, increasing context from $T=2$ to $T=4$ reduces MSE from $0.26$ to $0.16$ for camera motion and from $0.22$ to $0.19$ for object motion. 
CameraBench, whose camera-motion labels are clip-level, improves from $84.7\%$ to $85.3\%$. 
The effect is particularly strong on SSv2-$110$k, where accuracy increases monotonically from $16.7\%$ at $T=2$ to $23.1\%$ at $T=7$, beyond the training range of $T=5$. 
The multi-frame baseline based on AVG-pooled feature differences also benefits from additional context, showing that extra frames contain useful motion signal. However, it remains substantially below \method. The baseline with the maximum temporal context underperforms \method with a single frame pair ($T=2$) on all datasets shown. 
Thus, the gains of \method cannot be explained by access to additional frames alone.
Additional results in \Cref{sec:additional_temp_context_results_appendix} show that gains are less monotonic on datasets with more complex motion.
\begin{table*}[ht]
    \caption{\textbf{\method leads to structured motion representations where $p$ and $r$ specialize} in primary and residual motion when the features are used to linearly regress or classify motion or action. 
    We report MSE on standardized regression targets and accuracy in \% for classification tasks.
    Among individual tokens, the best result is shown in \textbf{bold}.
    Representations and targets that match are highlighted in \colorbox{matchcolour}{green}.
    The last row reports an upper-bound probe using the concatenated tokens $[p,r]$.
    }
    \label{tab:structured_motion}
    \centering
    \small
    \setlength{\tabcolsep}{0.4em}
    \begin{tabular}{l| ccccc c cc }
        \toprule
        \multirow{3}*{token} & \multicolumn{5}{c}{Regression (MSE $\downarrow$)} && \multicolumn{2}{c}{Classification (accuracy $\uparrow$)}\\
        \cmidrule{2-6} \cmidrule{8-9}
        & Kubric & Kubric & \multirow{2}*{DL3DV}& Static & Static && Camera & \multirow{2}*{SSv2-$110$k}\\
        & camera & object & & DAVIS2017 & YouTubeVOS && Bench & \\
        \midrule
        primary $p$ & \cellcolor{matchcolour}\textbf{0.16} & 0.21 & \cellcolor{matchcolour}\textbf{0.59} & \cellcolor{matchcolour}0.87 & \cellcolor{matchcolour}\textbf{0.71} && \cellcolor{matchcolour}\textbf{85.3} & \cellcolor{matchcolour}\textbf{23.1}\\
        residual $r$ & 0.28 & \cellcolor{matchcolour}\textbf{0.19} & 0.62 & \textbf{0.80} & 0.79 && 81.1 & 12.0 \\
        \midrule\midrule
        concat. $[p,r]$ & 0.16 & 0.14 & 0.56 & 0.81 & 0.71 && 88.0 & 22.1\\
        \bottomrule
    \end{tabular}
\end{table*}

\smallparagraph{Structured motion}
A key goal of \method is to produce \emph{structured} motion latents whose specialization follows the dominant motion regime. We test this by swapping which token is used for each probing task, and find that the learned roles support the primary/residual interpretation. Results are given in \Cref{tab:structured_motion}. 
On dynamic-camera datasets, $p$ is substantially better than $r$ for camera-motion prediction, while $r$ performs best for Kubric object motion, capturing dynamics left unexplained by the primary component. 
Under approximately static cameras, object motion becomes dominant, and $p$ performs best on static YouTubeVOS. Similarly, on SSv2-$110$k, where our camera-motion analysis indicates low inter-frame camera motion, $p$ performs best for action recognition.
Static DAVIS2017 is an exception with high variability across seeds for the residual token ($0.8\pm0.15$, see \Cref{sec:additional_structured_results_appendix}).
Concatenating the two tokens leads to similar performance, indicating that task-relevant information is primarily encoded in one of the two tokens. A notable exception is Kubric object motion, where predicting object translation in world coordinates can benefit from both the residual object-motion signal and the primary camera-motion component.

\subsection{Analysis and Ablation Studies}
\label{sec:ablation}

We summarize the main ablation findings below, with full component tables in Appendix \Cref{tab:ablation_appendix} and the backbone ablation in \Cref{tab:backbone_ablation}.

\smallparagraph{Weak anchors and sequential extraction promote specialization}
Removing either weak scene-level constraint weakens the primary/residual structure, while joint extraction makes the two tokens encode more overlapping motion.

\smallparagraph{Intermediate features and real data improve performance}
Intermediate encoder features outperform final-layer features alone and unlabeled real videos improve generalization beyond synthetic training. MSE improves Kubric object-motion probing significantly while $\mathcal{L}_1$ is slightly stronger on some real-video probes.

\smallparagraph{Static-camera regularization balances probes}
Varying $\lambda_{\mathrm{reg}}$ reveals a trade-off between Kubric object-motion probing and SSv2-$110$k action recognition. We use $\lambda_{\mathrm{reg}}=0.5$ as a balanced default.

\smallparagraph{\method works across pretrained backbones}
\Cref{tab:backbone_ablation} shows that \method can be applied to diverse image encoders. DINOv2/3-based backbones perform best overall.

\subsection{Qualitative evaluation}
We complement probing with qualitative analyses on different test samples, visualizing whether the two prediction stages explain complementary changes and whether motion tokens can be reused beyond the observed transition.

\begin{figure}[t]
  \centering
  \includegraphics[width=0.99\textwidth]{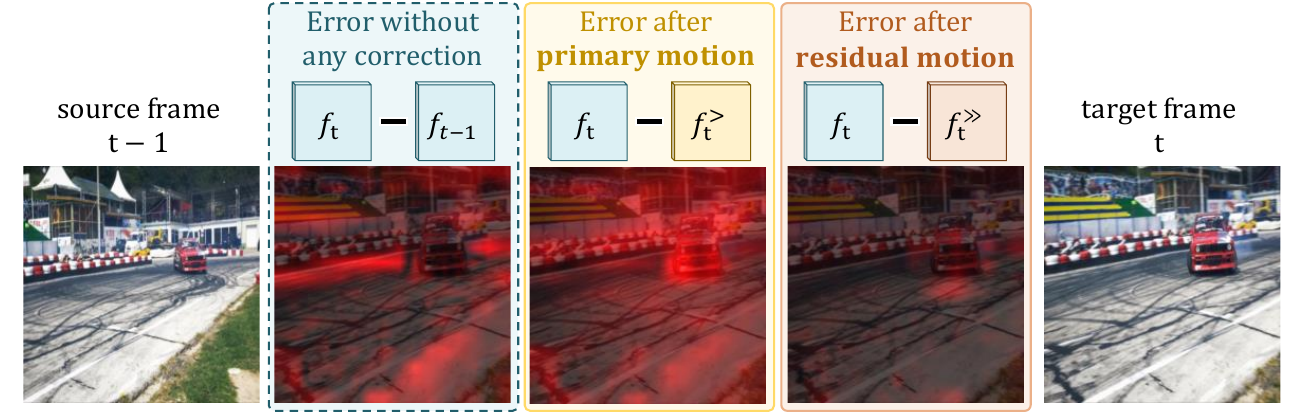}
  \caption{\textbf{Stage-wise structured feature prediction.}
    We visualize feature-map error to the target after primary and residual compensation on a sample from DAVIS2017. Errors are highlighted in red. 
    The \colorbox{prim}{\strut primary} stage corrects the dominant global change, including the camera-induced movement of the edges of the track. 
    Notably, primary motion compensation increases the error around the independently moving car.
    The \colorbox{res}{\strut residual} stage reduces this localized error.
  }
  \label{fig:motion_compensated_fmaps}
\end{figure}

\smallparagraph{Stage-wise feature compensation}
\Cref{fig:motion_compensated_fmaps} shows that the primary stage explains the dominant global transformation, while residual compensation reduces localized errors around the independently moving object.

\begin{figure}[ht]
  \centering
  \includegraphics[width=0.99\textwidth]{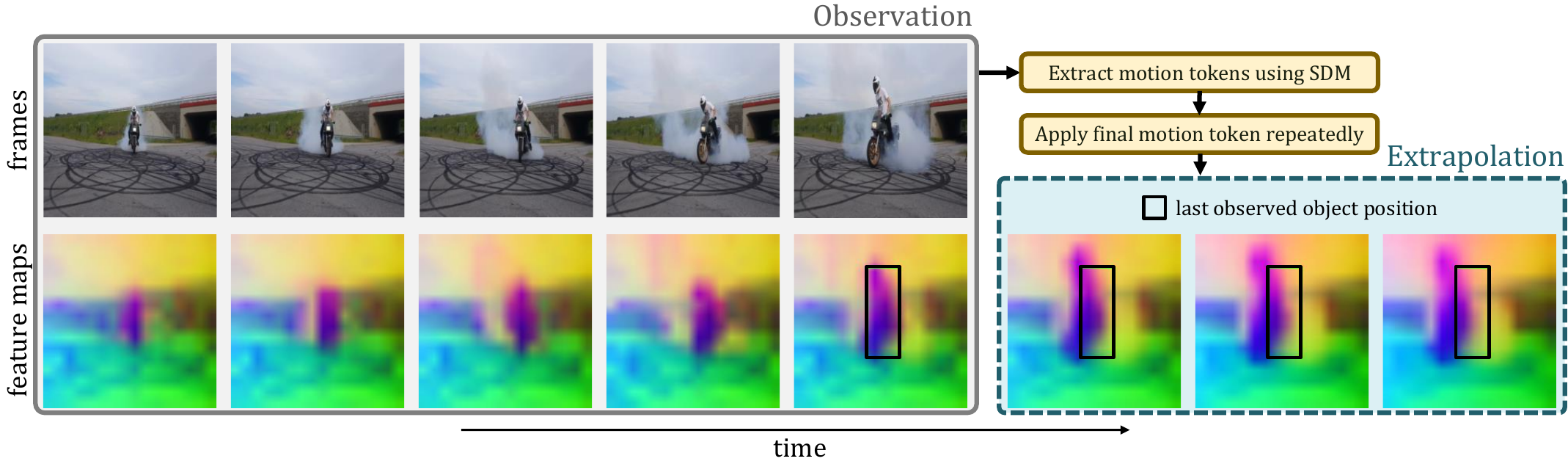}
  \caption{\textbf{Motion extrapolation.}
    We repeatedly apply the last motion token from the observed frames to predict future feature maps on a sample from the DAVIS2017 test split. Visualization is done by PCA projection to RGB after normalization. Black boxes mark the last observed ground-truth object position, which is repeated for the extrapolated feature maps. The extrapolated object features follow the object's motion beyond its last observed position over short horizons and drift at longer horizons.
  }
  \label{fig:motion_extrapolation}
\end{figure}

\smallparagraph{Motion extrapolation}
Repeatedly applying the last observed motion token yields plausible short-horizon feature extrapolations (\Cref{fig:motion_extrapolation}), but predictions drift over longer horizons. This suggests that the token stably captures local temporal dynamics, but does not support reliable long-range extrapolation without new observations.

\begin{figure}[ht]
  \centering
  \includegraphics[width=0.99\textwidth]{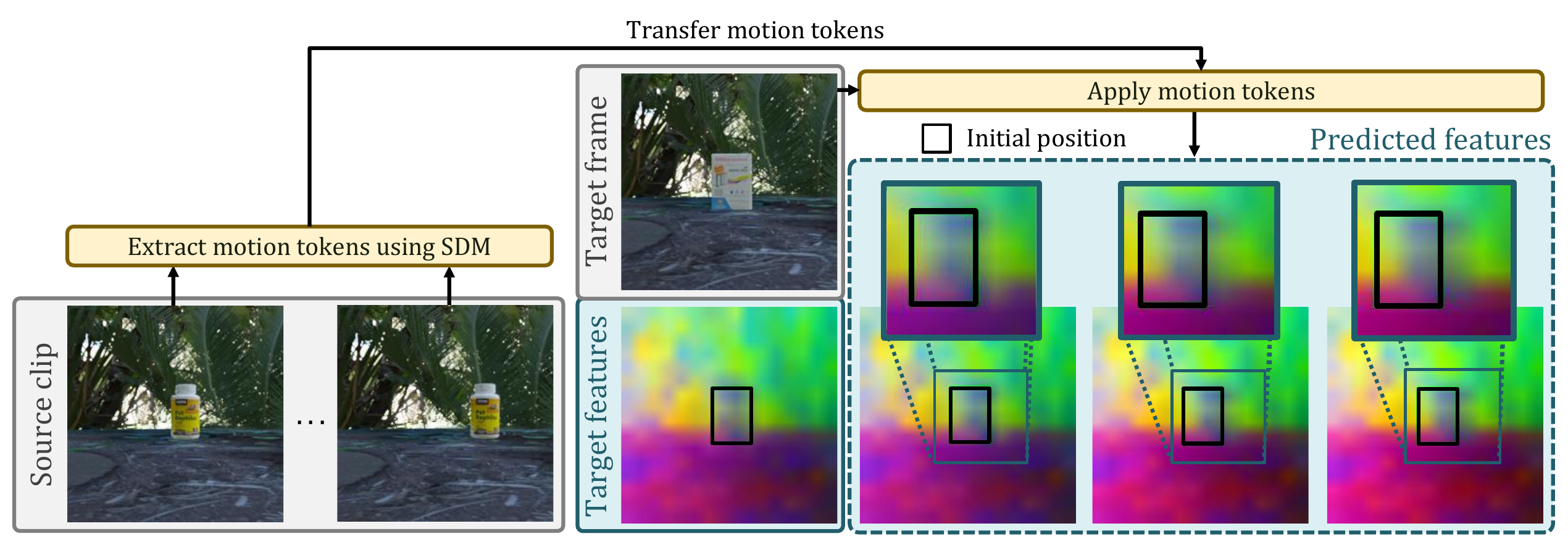}
  \caption{\textbf{Latent motion swapping.}
    We apply motion tokens from a source clip to the feature map of a target clip with a different object, both from the Kubric test set, and visualize the resulting features by PCA projection to RGB after normalization. Black boxes mark the initial object position. The target feature map follows the transferred motion over short horizons and drifts at longer horizons.
  }
  \label{fig:latent_swapping}
\end{figure}

\smallparagraph{Latent motion swapping}
Applying motion tokens from a source clip to a target clip with a different object induces consistent short-horizon feature-space motion (\Cref{fig:latent_swapping}). This indicates that the learned tokens can parameterize local and generalizable transformations across videos, although quality decreases over longer horizons.

%% file: sections/2_related_work.tex
\section{Related work}
\label{sec:related_work}
\textbf{Self-supervised image encoders.}
Self-supervised image encoders such as DINOv2~\cite{oquabdinov2}, AIM~\cite{el2024aim}, and MAE~\cite{he2022mae} provide strong frozen visual features. Although these image features contain useful 3D cues~\cite{el2024probe3d} and support recent reconstruction systems~\cite{wang2025vggt,lin2025depthanythingv3,wang2025pi3,luo20264rc}, they do not directly expose temporal or motion structure.

\textbf{Self-supervised video encoders.}
Self-supervised video encoders learn spatio-temporal representations directly from unlabeled videos, often through masked video modeling~\cite{tong2022videomae,wang2023videomaev2,carreira2024scaling4d,zoran2026rvm} or latent predictive objectives~\cite{bardes2024vjepa,assran2025vjepa2,mur2026vjepa21,wang2025poodle}. These methods learn general video representations, whereas \method keeps an image backbone frozen and learns explicit primary/residual motion tokens on top.

\textbf{Training image encoders on videos.}
Several works adapt image-pretrained models with video data, for example through temporal feature consistency~\cite{salehi2025mosic,salehi2023timedoestell} or correspondence learning in SiamMAE~\cite{gupta2023siammae}. Others show that long, uncurated videos can train strong image encoders from scratch~\cite{venkataramanan2024imagenetonevideo}. We instead keep the image backbone frozen and learn structured motion tokens on top.

\textbf{3D reconstruction models.}
Models such as DUSt3R~\cite{wang2024dust3r}, VGGT~\cite{wang2025vggt}, and CUT3R~\cite{wang2025cut3r} are purpose-built for 3D reconstruction, estimating camera pose, depth, point maps, and related geometry from strong 3D supervision. Recent extensions target dynamic scenes and 4D reconstruction~\cite{zhangmonst3r,hu2025vggt4d}. In contrast, \method does not reconstruct full 3D geometry, but uses much weaker supervision to remodel frozen image features into primary/residual motion tokens.

\textbf{Latent action and world models.}
Latent action and world models learn from video by predicting future frames or pretrained feature representations, avoiding dense annotations and, in feature-space methods, pixel-level generation~\cite{kerssies2026frame,baldassarre2025backtothefeatures,garrido2026latentinthewild}. Prior work often represents temporal change with spatial future tokens or compact latent actions~\cite{baldassarre2025backtothefeatures,garrido2026latentinthewild,kerssies2026frame,boduljak2025vfmf,zhou2025dinoworld}. \method is closest to feature-space future prediction, but explicitly structures motion into primary and residual components rather than encoding change in a single latent action.

%% file: sections/5_conclusion.tex
\section{Conclusion}
\label{sec:conclusion}

We studied whether structured motion representations could be extracted from frozen pretrained image backbones and proposed the Structured Dynamics Model (SDM). SDM learns structured primary/residual motion latents through future-feature prediction, using weak scene-level supervision on synthetic videos together with unlabeled real data. 
Our results suggest that frozen image backbones can be remodeled into structured video-dynamics representations, enabling interpretable temporal abstractions to emerge without full video pretraining while remaining competitive with substantially larger supervised geometry models.

The learned tokens specialize according to scene dynamics. The primary token captures camera motion in dynamic-camera videos and object motion in static-camera videos, while the residual token captures remaining object-centric dynamics when primary motion explains camera-induced change.
This mirrors the effectiveness of iterative coarse-to-fine refinement in other domains, from optical flow estimation to diffusion models, and illustrates that residual compensation is a useful inductive bias.

%% file: sections/X_suppl.tex
\section{Dataset and Evaluation Details}\label{sec:data_eval_details_appendix}

\begin{table*}[ht]
    \caption{\textbf{Evaluation tasks.}
    Summary of the probing tasks used for each evaluation dataset.}
    \label{tab:evaluation_tasks}
    \centering
    \small

    \begin{tabular}{l | ccc}
        \toprule
        & Kubric & DL3DV & CameraBench\\
        \midrule
        \subtitlerow{4}{Task}
        target & camera / object motion & camera motion & camera-motion class\\
        target format & 6-DoF pose / 3D translation & 6-DoF pose & discrete motion label\\
        probe type & regression & regression & classification\\
        metric & norm. MSE & norm. MSE & top-1 acc.\\
        \midrule
        \subtitlerow{4}{Representation}
        probed token & $p$ / $r$ & $p$ & $p$\\
        temporal target & frame-pair motion & frame-pair motion & clip-level motion\\
        \midrule
        \subtitlerow{4}{Split and sampling}
        split & generated train/test & video-level 70/30 & video-level 70/30\\
        train sequences / video & 1 & 3 & 1\\
        test sequences / video & 1 & 3 & 1\\
        balancing & motion-mag. subsampling & motion-mag. subsampling & class downsampling\\
        \midrule\midrule
        & Static DAVIS2017 & Static YouTubeVOS & SSv2-$110$k\\
        \midrule
        \subtitlerow{4}{Task}
        target & object displacement & object displacement & action class\\
        target format & 2D displacement & 2D displacement & action label\\
        probe type & regression & regression & classification\\
        metric & norm. MSE & norm. MSE & top-1 acc.\\
        \midrule
        \subtitlerow{4}{Representation}
        probed token & $p$ & $p$ & $p$\\
        temporal target & frame-pair motion & frame-pair motion & clip label\\
        \midrule
        \subtitlerow{4}{Split and sampling}
        split & official train/test & video-level 70/30 & official train/val\\
        train sequences / video & 3 & 3 & 1\\
        test sequences / video & 3 & 3 & 1\\
        balancing & motion-mag. subsampling & motion-mag. subsampling & class downsampling\\
        \bottomrule
    \end{tabular}
\end{table*}
For regression tasks, motion-magnitude balancing denotes retaining high-motion samples and subsampling low-motion samples as described in \Cref{sec:data_split_sampling_appendix}.

\subsection{Kubric generation}
\label{sec:kubric_details_appendix}

\begin{figure}[ht]
  \centering

  \begin{subfigure}{0.95\linewidth}
    \centering
    \includegraphics[width=\linewidth]{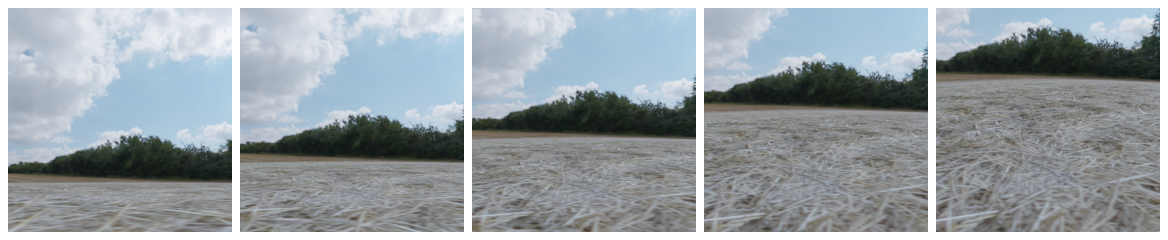}
    \caption{Moving camera, static scene.}
    \label{fig:kubric_example_moving_camera_static_scene}
  \end{subfigure}

  \vspace{2mm}

  \begin{subfigure}{0.95\linewidth}
    \centering
    \includegraphics[width=\linewidth]{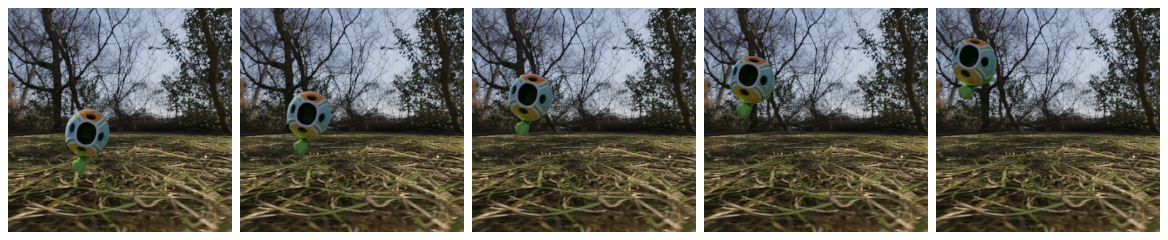}
    \caption{Static camera, dynamic scene.}
    \label{fig:kubric_example_static_camera_dynamic_scene}
  \end{subfigure}

  \vspace{2mm}

  \begin{subfigure}{0.95\linewidth}
    \centering
    \includegraphics[width=\linewidth]{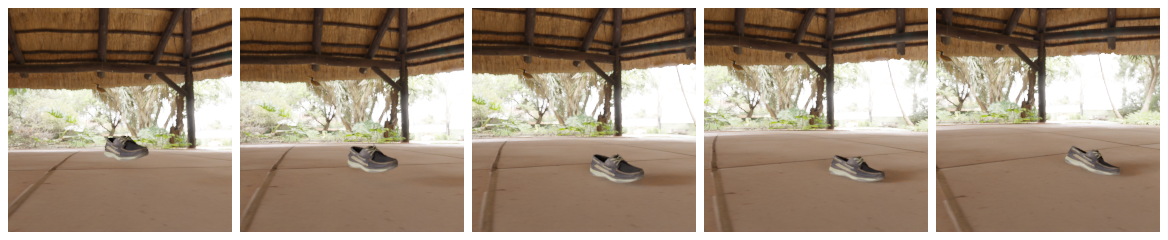}
    \caption{Moving camera, dynamic scene.}
    \label{fig:kubric_example_moving_camera_dynamic_scene}
  \end{subfigure}

  \caption{\textbf{Kubric examples.} Example sequences from our generated Kubric training data. The three subsets provide weak scene-level supervision for disentangling primary and residual dynamics: moving-camera/static-scene sequences contain only camera-induced motion, static-camera/dynamic-scene sequences contain only object motion, and moving-camera/dynamic-scene sequences combine both sources of motion. Examples are shown at $1$ fps.}
  \label{fig:kubric_example}
\end{figure}

We generate our synthetic training and evaluation data with Kubric~\cite{greff2021kubric}, starting from the MOVi-E setup\footnote{https://github.com/google-research/kubric/tree/main/challenges/movi}.
Each sample is generated from a unique random seed. Additionally, training and evaluation samples use disjoint subsets of objects and backgrounds.
We render short clips of $10$ frames at $2$ fps and resolution $224\times224$.
We use three scene configurations: static scene with moving camera, dynamic scene with static camera, and dynamic scene with moving camera.
Static scenes contain no foreground objects, whereas dynamic scenes contain a single moving object.

For dynamic scenes, we disable gravity and assign the object a randomly sampled initial velocity.
This produces approximately unbiased object motions that are not dominated by falling trajectories.
For camera-dynamic scenes, the camera follows a linear trajectory while continuously looking at a fixed point above the world origin.
This yields controlled ego-motion while keeping the scene center approximately in view.

We apply several pre-rendering validity checks to ensure that dynamic objects are sufficiently visible.
First, we require the selected object to have a minimum size of one simulator unit, measured as the shortest side of its 3D bounding box.
If the sampled object is too small, we resample the object. If no valid object is found after $100$ attempts, the sample is rejected.
Second, for all frames, we require at least $50\%$ of the projected 2D bounding box of the object to lie inside the image.
Third, we reject samples in which the object lies behind the camera, i.e., when its maximum depth satisfies \texttt{zmax}$<0$.
These checks are performed before rendering for efficiency. Invalid samples are discarded and regenerated with a new seed.

Because of the low resolution, generation can be performed without a GPU and takes approximately $12$ seconds per valid sample. Invalid samples take only $0.3$ seconds due to the pre-rendering checks. Approximately $90\%$ of samples with dynamic objects pass the validity checks.

\subsection{Static-camera object-motion benchmarks}\label{sec:static_davis_youtubevos_construction_appendix}
\textbf{Inputs.}
We build a real-world object-motion benchmark with approximately static camera from DAVIS2017 and YouTubeVOS.
We start from videos that contain both RGB frames and object annotations, and keep only timestamps for which both modalities are available.
For each scene, we summarize the annotation masks by recording per-frame visible object area and image-plane centroid.

\textbf{Camera-motion filtering.}
To identify near-static-camera clips, we estimate camera motion from sampled video frames using VGGT-based camera-pose prediction, which provides per-interval camera rotation and translation estimates.
We then construct candidate windows of length $T{=}4$ at $2$ fps and retain only those windows whose maximum local camera motion across consecutive frame pairs stays below fixed translation and rotation thresholds. Specifically, we require the maximum relative rotation between consecutive frames to be below $2^\circ$ and the maximum translation between camera centers to be below $0.05$.

\textbf{Object filtering.}
Among the near-static windows, we further keep only samples containing sufficiently large objects, requiring objects to span more than $15$ pixels and to move by at least $0.05$ in normalized centroid displacement.

\textbf{Motion targets.}
For each retained window, we compute continuous object motion from the selected object's mask sequence.
We extract the object centroid in each frame and define motion as the frame-to-frame centroid displacement in image coordinates.
These displacements are normalized by image width and height, yielding continuous horizontal and vertical motion components in relative image-plane units.

\subsection{Dataset splits and sampling}\label{sec:data_split_sampling_appendix}
\textbf{Dataset-specific probing splits.}
SSv2 uses the official train/validation split filtered by videos that are too short for the required clip sampling.
CameraBench is an evaluation benchmark and does not provide a separate training split. We therefore first balance each selected class group and then form a pseudo split with a $70/30$ train/test ratio.
For DL3DV, we use the \texttt{1K} to \texttt{7K} subsets and form splits at the video level to avoid leakage, assigning $70\%$ of videos to training and $30\%$ to testing. We do not use the official test split because it is too small for our probing setting. 
To further increase the number of probing samples, we sample three sequences per video for both training and testing. Videos in the probing test split are excluded from unlabeled \method training.
For static YouTubeVOS, we also use a video-level pseudo split because the official test split does not provide object segmentations at sufficient temporal resolution. Similar to DL3DV, we sample three sequences per training and test video.
For static DAVIS2017, we use the official train/test split and also sample three sequences per video.
For the static-camera object-motion probes, we train on the combined static DAVIS2017 and static YouTubeVOS training sets to increase probe-training data, while reporting evaluation results separately on each benchmark.
Kubric uses predefined train and test sets for the moving-camera/dynamic-scene setting. The actual probing splits are further balanced by regression bins, with bin edges computed on the training split and reused on the test split.

\textbf{Balancing.}
All classification probing datasets are class-balanced by downsampling each class to the smallest class count. 
For regression probing, we downsample low-motion samples by constructing a tail-aware subset. 
We retain all samples above the 75th percentile of motion magnitude and randomly subsample the remaining lower-motion samples to the same count. 
This yields equal numbers of high- and lower-motion samples and prevents regression probes from being dominated by near-static examples.

\subsection{Probe targets}\label{sec:probe_targets_appendix}
\textbf{Regression tasks.}
For regression, we train one linear probe per probing setting, where each probe jointly predicts the full target vector for that setting.
Camera-motion regression predicts the 6D vector $[v_x, v_y, v_z, \mathrm{yaw}, \mathrm{pitch}, \mathrm{roll}]$.
This is used for DL3DV and Kubric camera-motion evaluation.
Kubric object-motion regression predicts the 3D translation vector $[v_x, v_y, v_z]$, since objects do not rotate in our Kubric setup.
The relative pose is computed between frame $t{+}1$ and frame $t$, expressed in the coordinate system of frame $t{+}1$.
Static DAVIS2017 and static YouTubeVOS object-motion regression predicts the 2D image-plane centroid displacement $[\Delta x, \Delta y]$ in normalized image coordinates.

\textbf{Classification tasks.}
For classification, we train one linear probe for each selected label set.
SSv2-$110$k uses full-class action recognition.
CameraBench uses separate probes for eight camera-motion groups: speed (\texttt{slow}, \texttt{regular}, \texttt{fast}), dolly (\texttt{in}, \texttt{out}), pedestal (\texttt{up}, \texttt{down}), truck (\texttt{left}, \texttt{right}), pan (\texttt{left}, \texttt{right}), tilt (\texttt{up}, \texttt{down}), roll (\texttt{CW}, \texttt{CCW}), and zoom (\texttt{in}, \texttt{out}).

\subsection{Probe training}\label{sec:probe_training_appendix}
Our evaluation uses linear probes trained on frozen motion representations.
Each probe consists of a single linear layer on top of the corresponding motion token and is trained for $100$ epochs with SGD, momentum $0.9$, cosine learning-rate decay, and evaluation after every epoch.
Following prior work~\cite{oquabdinov2}, we train probes with multiple learning rates in parallel. Specifically we use:
\begin{equation}
    \mathrm{lr} \in \{1,2,5\}\times 10^{\{-5,-4,-3,-2,-1\}} \cup \{3{\times}10^{-1}\},
\end{equation}
which corresponds to the union of the learning rates used in the DINOv2 paper and its official implementation.
We linearly scale these learning rates from a reference batch size of $256$ to batch size $16$ for CameraBench and batch size $64$ for all other datasets.
For regression probes, we additionally apply gradient clipping with norm $5$ to improve optimization stability.
Depending on the task, we probe either the primary motion token $p$ or the residual motion token $r$. The choice is stated in the corresponding result table.

As we train many probes, we improve the efficiency of probe training by using no input augmentations and precomputing the motion features before training the linear probes.

\subsection{Metrics and probe selection}\label{sec:metrics_probe_selection_appendix}
For classification, we report the top-1 accuracy averaged over the different subtasks.
For regression, we report mean squared error on targets normalized with the training-set mean and standard deviation.
Throughout, we report the best held-out metric over probe training and across the parallel learning-rate sweep rather than the last-epoch metric of a single probe.
This reduces sensitivity to the exact probe optimization hyperparameters when comparing many probes across datasets, targets, and representations. 
All methods and baselines use the same protocol. 

\subsection{DeltaTok evaluation}
\label{sec:deltatok_eval_appendix}
We evaluate DeltaTok~\cite{kerssies2026frame} as a strong self-supervised baseline using the released Kinetics-700~\cite{carreira2019kinetics700} ViT-B model and the same evaluation protocol as for \method.
We follow its inference protocol by resizing frames to $512{\times}512$ and extracting the final delta token for each clip as its motion descriptor.

DeltaTok and \method are not trained under matched compute or input-resolution budgets.
DeltaTok uses $512{\times}512$ inputs, DINOv3~\cite{simeoni2025dinov3} features, and is trained for $1.6$M iterations with an effective batch size of $1024$.
In comparison, \method uses $224{\times}224$ inputs, DINOv2-with-registers~\cite{darcet2024dino_registers} features by default, and is trained for $200$k iterations with batch size $128$.
Thus, DeltaTok uses $8{\times}$ more optimization iterations and an $8{\times}$ larger effective batch size, in addition to operating at substantially higher spatial resolution.

\subsection{Supervised model evaluation}
\label{sec:supervised_model_eval_appendix}
We evaluate supervised 3D and geometry-oriented models as frozen feature extractors. For each model, we use its default inference preprocessing, run inference on the same frame sequences as in our probing datasets, and train linear probes using the same protocol as for \method. We derive motion descriptors from consecutive frames using the same temporal concatenation or difference baselines used for frozen image features.

For VGGT, we use \texttt{VGGT-1B}. Frames are converted to RGB, resized to a target width $518$ while preserving aspect ratio, rounded to a height divisible by $14$, center-cropped to $518{\times}518$, and scaled to $[0,1]$. We use either the camera token or mean-pooled patch tokens from the final aggregated token map. 
For Depth Anything 3, we use \texttt{DA3-GIANT-1.1} with the official inference pipeline and extract either camera-token or mean-pooled patch-token features from the final backbone output. 
For Pi3X, we resize RGB inputs with preserved aspect ratio to the largest resolution with area at most $255{,}000$ pixels and dimensions divisible by $14$, and extract the output features of the decoder. The final descriptor is obtained either by averaging the features of the patch tokens or, as Pi3X does not have a dedicated camera token, by averaging the $5$ registers.

\subsection{Sample visualization details}
\label{sec:sample_vis_details_appendix}
For qualitative sample visualization, we project dense feature maps to RGB using a sample-specific PCA. For a given sample, we collect the feature tokens from all frames of the sequences being compared, flatten them into a single feature matrix, and normalize them. The PCA is fit jointly on all extracted feature maps.

We retain the first three principal components and map them to the three RGB channels. After projection, each component is scaled independently to $[0,1]$ using the 1st and 99th percentiles computed on the PCA fit set, and the resulting token grid is reshaped to the spatial feature-map layout and bilinearly upsampled to the image resolution for display.

Pixel-wise error maps are the L2 distance between predicted and ground-truth feature maps, computed in the projected space. These distances are normalized with a shared 99th-percentile scale across all compared maps, converted to a heatmap, and overlaid on the corresponding RGB video frames.

\section{Additional results}\label{sec:additional_results_appendix}
\subsection{SSv2 camera motion analysis}\label{sec:ssv2_camera_motion_appendix}
To motivate the choice of token to use for SSv2-$110$k action recognition probing, we analyze the amount of camera motion in SSv2 clips.
We randomly sample $5$k SSv2 clips, estimate frame-to-frame camera motion using VGGT~\cite{wang2025vggt} and compute the average per-step camera rotation and translation.
To contextualize these values, we apply the same procedure to the CameraBench dataset and aggregate the camera motion separately for samples annotated as \texttt{slow-speed}, \texttt{regular-speed}, or \texttt{fast-speed}.
The \texttt{slow-speed} subset includes not only clips with little camera motion but also all samples with a static camera.

\begin{table}[h]
    \caption{\textbf{Camera-motion statistics for SSv2 and CameraBench.}
    We report average frame-to-frame camera motion estimated with VGGT. Rotation is reported in degrees per second while translation is in VGGT's normalized scene-scale units per second.
    SSv2 exhibits lower camera motion than regular- and fast-speed CameraBench clips and is closer to the slow-speed subset in rotation.}
    \label{tab:ssv2_camera_motion}
    \centering
    \small
    \begin{tabular}{l | cccc}
        \toprule
        & \multirow{2}*{SSv2} & \multicolumn{3}{c}{CameraBench}\\
        & & \texttt{slow-speed} & \texttt{regular-speed} & \texttt{fast-speed} \\
        \midrule
        rotation & $3.89$ & $1.57$ & $7.77$ & $32.89$ \\
        translation & $0.08$ & $0.03$ & $0.13$ & $0.62$ \\
        \bottomrule
    \end{tabular}
\end{table}

SSv2 camera motion lies between the slow- and regular-speed CameraBench subsets and is closer to the slow-speed subset in terms of rotation.
This suggests that many SSv2 clips contain limited inter-frame camera motion.
We therefore follow a similar reasoning as for the static-camera DAVIS2017 and YouTubeVOS object-motion benchmarks (see \Cref{sec:training_eval_details}). When camera-induced change is small, action-related object dynamics are expected to be a dominant source of temporal change, and we probe the primary token $p$ for SSv2 action recognition.

\subsection{Baseline results}\label{sec:baseline_results_appendix}
We compare different ways to derive motion descriptors from strongly supervised and baseline models in \Cref{tab:baseline_descriptor_ablation,tab:multi_layer_baseline_ablation_appendix}.
We use these ablations to choose between concatenation and feature differences for the direct-feature baselines and supervised-model descriptors reported in the main paper.
For single-layer DINOv2 features, the best descriptor depends on the pooling strategy and task. Feature differences of AVG-pooled features perform best overall, especially on regression tasks, while concatenating \cls{} tokens of consecutive frames works best for classification tasks.
For the supervised models, feature-difference descriptors tend to perform best across all benchmarks except SSv2-$110$k. On this action recognition benchmark, all three supervised models show the same pattern. Feature differences perform poorly, with the best accuracy reaching only $6.2\%$, whereas concatenation-based descriptors achieve between $12.2\%$ and $19.8\%$. 
A possible explanation is that, unlike the other benchmarks, SSv2-$110$k uses action labels rather than motion targets, and these labels may depend on semantic information that is reduced when subtracting features.
Finally, \Cref{tab:multi_layer_baseline_ablation_appendix} shows that giving direct \cls{} and AVG-pool baselines access to all $12$ DINOv2 blocks improves these baselines but does not close the gap to \method.
Thus, the improvements of \method are not explained by access to intermediate features alone, but by using structured recurrent prediction to turn these features into motion representations.
\begin{table*}[ht]
    \caption{\textbf{Baseline descriptor ablations.}
    We compare descriptor choices for frozen DINOv2-with-registers representations and strongly supervised representations. 
    For DINOv2 with registers, we evaluate \cls{} tokens and average-pooled patch features using either concatenation or feature differences across consecutive frames. Multi-frame variants average frame-pair descriptors over time. 
    For VGGT and Depth Anything 3, we compare average-pooled feature maps and camera tokens using either concatenation or differences across consecutive frames. 
    Since Pi3X does not have a camera token, we use the average of its registers as an alternative and otherwise follow the same setup.
    We report MSE on standardized regression targets and accuracy in \% for classification tasks. The best results for each model and baseline type are highlighted in \textbf{bold}.
    }
    \label{tab:baseline_descriptor_ablation}
    \centering
    \footnotesize
    \setlength{\tabcolsep}{0.35em}
    \begin{tabular}{l ccccc c cc}
        \toprule
        \multirow{3}*{Method} & \multicolumn{5}{c}{Regression (MSE $\downarrow$)} && \multicolumn{2}{c}{Classification (accuracy $\uparrow$)}\\
        \cmidrule{2-6} \cmidrule{8-9}
        & \multicolumn{2}{c}{Kubric} & \multirow{2}*{DL3DV} & Static & Static && Camera & \multirow{2}*{SSv2-$110$k}\\
        & cam & obj & & DAVIS2017 & YouTubeVOS && Bench & \\
        \midrule
        \subtitlerow{9}{DINOv2 with registers baselines}
        AVG-pool concat & 1.05\std{0.0} & 0.73\std{0.0} & 1.06\std{0.0} & 0.96\std{0.03} & 0.97\std{0.0} && 68.4\std{1.3} & \textbf{14.1}\std{0.1}\\
        AVG-pool concat, multi-frame & 0.97\std{0.0} & \textbf{0.69}\std{0.0} & 1.07\std{0.0} & 0.96\std{0.02} & 0.95\std{0.01} && 67.5\std{1.2} & 14.0\std{0.1}\\
        AVG-pool diff & 0.96\std{0.0} & 0.93\std{0.0} & \textbf{1.04}\std{0.0} & \textbf{0.78}\std{0.02} & \textbf{0.91}\std{0.0} && 67.0\std{1.2} & 2.3\std{0.0}\\
        AVG-pool diff, multi-frame & \textbf{0.87}\std{0.0} & 0.72\std{0.0} & 1.05\std{0.0} & 1.02\std{0.00} & 0.92\std{0.0} && \textbf{69.2}\std{0.9} & 13.3\std{0.0}\\
        \midrule
        \cls{} concat & 1.16\std{0.0} & 0.78\std{0.0} & 1.09\std{0.0} & \textbf{0.90}\std{0.10} & 1.05\std{0.01} && 68.6\std{0.8} & 14.9\std{0.2}\\
        \cls{} concat, multi-frame & 1.05\std{0.0} & \textbf{0.77}\std{0.0} & 1.09\std{0.0} & 0.92\std{0.02} & 0.98\std{0.0} && \textbf{68.7}\std{1.1} & \textbf{15.5}\std{0.1}\\
        \cls{} diff & 1.09\std{0.0} & 0.96\std{0.0} & 1.09\std{0.0} & \textbf{0.90}\std{0.01} & \textbf{0.93}\std{0.0} && 67.1\std{1.0} & 3.0\std{0.0}\\
        \cls{} diff, multi-frame & \textbf{0.96}\std{0.0} & 0.80\std{0.0} & \textbf{1.07}\std{0.0} & 1.01\std{0.01} & \textbf{0.93}\std{0.0} && 67.3\std{0.6} & 12.5\std{0.1}\\
        \midrule
        \subtitlerow{9}{strongly supervised}
        VGGT AVG feat. map concat & 0.18\std{0.0} & 0.42\std{0.0} & 0.59\std{0.0} & 0.99\std{0.06} & 0.95\std{0.0} && 82.9\std{0.0} & \textbf{15.5}\std{0.1}\\
        VGGT AVG feat. map diff & 0.09\std{0.0} & \textbf{0.32}\std{0.0} & 0.56\std{0.0} & 0.78\std{0.10} & \textbf{0.79}\std{0.0} && 82.4\std{0.3} & 5.6\std{0.0}\\
        VGGT cam. token concat & 0.05\std{0.0} & 0.43\std{0.0} & \textbf{0.53}\std{0.0} & 0.97\std{0.07} & 0.98\std{0.0} && 87.5\std{0.3} & 13.5\std{0.1}\\
        VGGT cam. token diff & \textbf{0.03}\std{0.0} & 0.44\std{0.0} & \textbf{0.53}\std{0.0} & \textbf{0.71}\std{0.04} & 0.81\std{0.0} && \textbf{88.7}\std{0.7} & 4.8\std{0.0}\\
        \midrule
        DA3 AVG feat. map concat & 0.28\std{0.0} & 0.52\std{0.01} & 0.66\std{0.01} & 1.89\std{0.20} & 1.65\std{0.02} && 83.3\std{0.4} & \textbf{14.0}\std{0.1}\\
        DA3 AVG feat. map diff & 0.11\std{0.0} & \textbf{0.40}\std{0.0} & 0.55\std{0.0} & \textbf{0.87}\std{0.11} & \textbf{0.89}\std{0.01} && 82.9\std{1.1} & 5.2\std{0.1}\\
        DA3 cam. token concat & 0.18\std{0.0} & 0.72\std{0.01} & 0.62\std{0.0} & 3.53\std{1.04} & 1.66\std{0.06} && 81.5\std{0.2} & 12.2\std{0.4}\\
        DA3 cam. token diff & \textbf{0.04}\std{0.0} & 0.58\std{0.0} & \textbf{0.53}\std{0.0} & \textbf{0.87}\std{0.04} & 0.91\std{0.02} && \textbf{84.4}\std{1.3} & 4.6\std{0.1}\\
        \midrule
        Pi3X AVG feat. map concat & 0.10\std{0.0} & 0.44\std{0.0} & 0.54\std{0.0} & 0.71\std{0.01} & 0.86\std{0.0} && \textbf{87.9}\std{0.7} & \textbf{19.8}\std{0.1}\\
        Pi3X AVG feat. map diff & \textbf{0.07}\std{0.0} & \textbf{0.38}\std{0.00} & \textbf{0.52}\std{0.0} & \textbf{0.52}\std{0.00} & \textbf{0.75}\std{0.0} && 86.8\std{0.8} & 6.2\std{0.1}\\
        Pi3X avg. registers concat & 0.33\std{0.0} & 0.77\std{0.0} & 0.65\std{0.0} & 0.95\std{0.03} & 0.94\std{0.0} && 79.8\std{1.6} & 12.7\std{0.3}\\
        Pi3X avg. registers diff & 0.08\std{0.0} & 0.45\std{0.0} & 0.53\std{0.0} & 0.61\std{0.01} & 0.77\std{0.0} && 85.7\std{0.4} & 5.7\std{0.1}\\
        \midrule
        \rowcolor{methodblue}
        \method & {0.16}\std{0.0} & {0.19}\std{0.0} & {0.59}\std{0.0} & {0.87}\std{0.1} & {0.71}\std{0.0} && {85.3}\std{1.7} & {23.1}\std{1.3}\\
        \bottomrule
    \end{tabular}
\end{table*}
\begin{table*}[ht]
    \caption{\textbf{Multi-layer direct-feature baselines.}
        To control for the use of intermediate DINOv2 features, we evaluate direct \cls and average-pooled patch descriptors extracted from all $12$ backbone blocks. Descriptors are averaged across layers and combined across adjacent frames either by concatenation or feature difference. 
        Even with access to multi-layer features, direct descriptors remain substantially below \method on all datasets except for static DAVIS2017, which is consistent with the results obtained without multi-layer features (see \Cref{tab:baselines}). This indicates that the gains are not explained by intermediate-feature aggregation alone. 
        We report MSE on standardized regression targets and accuracy in \% for classification tasks.
        Best results are highlighted in \textbf{bold}.
    }
    \label{tab:multi_layer_baseline_ablation_appendix}
    \centering
    \footnotesize
    \setlength{\tabcolsep}{0.35em}
    \begin{tabular}{l ccccc c cc}
        \toprule
        \multirow{3}*{Method} & \multicolumn{5}{c}{Regression (MSE $\downarrow$)} && \multicolumn{2}{c}{Classification (accuracy $\uparrow$)}\\
        \cmidrule{2-6} \cmidrule{8-9}
        & \multicolumn{2}{c}{Kubric} & \multirow{2}*{DL3DV} & Static & Static && Camera & \multirow{2}*{SSv2-$110$k}\\
        & cam & obj & & DAVIS2017 & YouTubeVOS && Bench & \\
        \midrule
        \subtitlerow{9}{DINOv2 with registers, multi-layer baselines}
        AVG-pool concat & 0.94\std{0.0} & 0.66\std{0.0} & 1.02\std{0.0} & 0.81\std{0.01} & 0.94\std{0.0} && 67.0\std{1.7} & 10.6\std{0.2}\\
        AVG-pool concat, multi-frame & 0.91\std{0.0} & 0.70\std{0.0} & 1.04\std{0.0} & 0.89\std{0.02} & 0.95\std{0.0} && 66.8\std{0.9} & 12.6\std{0.1}\\
        AVG-pool diff & 0.88\std{0.0} & 0.83\std{0.0} & 1.04\std{0.0} & \textbf{0.65}\std{0.02} & 0.83\std{0.0} && 65.5\std{0.9} & 1.9\std{0.0}\\
        AVG-pool diff, multi-frame & 0.82\std{0.0} & 0.60\std{0.0} & 1.06\std{0.0} & 1.01\std{0.01} & 0.91\std{0.0} && 70.0\std{0.6} & 6.9\std{0.0}\\
        \midrule
        \cls{} concat & 1.05\std{0.0} & 0.72\std{0.0} & 1.05\std{0.0} & 0.93\std{0.02} & 0.94\std{0.01} && 68.1\std{0.5} & 13.6\std{0.3}\\
        \cls{} concat, multi-frame & 1.01\std{0.0} & 0.77\std{0.0} & 1.06\std{0.0} & 0.94\std{0.0} & 0.94\std{0.0} && 67.9\std{0.9} & 14.5\std{0.4}\\
        \cls{} diff & 0.98\std{0.0} & 0.91\std{0.0} & 1.04\std{0.0} & 1.00\std{0.01} & 0.92\std{0.0} && 65.8\std{3.3} & 2.0\std{0.1}\\
        \cls{} diff, multi-frame & 0.88\std{0.0} & 0.68\std{0.0} & 1.05\std{0.0} & 1.02\std{0.0} & 0.93\std{0.0} && 68.6\std{2.0} & 7.5\std{0.0}\\
        \midrule
        \rowcolor{methodblue}
        \method & \textbf{0.16}\std{0.0} & \textbf{0.19}\std{0.0} & \textbf{0.59}\std{0.0} & {0.87}\std{0.1} & \textbf{0.71}\std{0.0} && \textbf{85.3}\std{1.7} & \textbf{23.1}\std{1.3}\\
        \bottomrule
    \end{tabular}
\end{table*}

\subsection{Temporal context}\label{sec:additional_temp_context_results_appendix}
\Cref{tab:temporal_context_regression_appendix,tab:temporal_context_ssv2_appendix} provide the full temporal-context results with standard deviations. 
\method benefits most from longer context on temporally coherent benchmarks, such as linear Kubric motion and SSv2-$110$k, where performance improves monotonically even beyond $5$ frames, the longest context used during training. 
The same trend holds for the multi-frame baseline based on average-pooled feature differences, which nevertheless remains below \method even when using longer context.
In contrast, \method remains stable on datasets with less constrained motion, such as DL3DV, static YouTubeVOS, and static DAVIS2017, where the baseline's performance even deteriorates significantly with increasing temporal context.
Overall, additional context is useful when motion is coherent, but not uniformly beneficial for unconstrained real-world videos.

\begin{table*}[ht]
    \caption{\textbf{Additional temporal-context results.} \method and the baseline generally benefit from additional frames on the temporally coherent Kubric and CameraBench benchmarks.
    On benchmarks with less temporal coherence, our method remains stable with increasing temporal context, while the frame-pair baseline can degrade.
    We report MSE on standardized regression targets and accuracy in \% for classification tasks.
    }
    \label{tab:temporal_context_regression_appendix}
    \centering
    \small
    \setlength{\tabcolsep}{0.4em}
    \begin{tabular}{l| cc | cc | cc }
        \toprule
        \multirow{2}*{method} & \multicolumn{2}{c}{Kubric camera $\downarrow$}&\multicolumn{2}{c}{Kubric object $\downarrow$} & \multicolumn{2}{c}{DL3DV $\downarrow$}\\
        & T=2 & T=4 & T=2 & T=4& T=2 & T=5\\
        \midrule
        baseline & 0.96\std{0.0} & 0.87\std{0.0}\gooddelta{-0.09} & 0.93\std{0.0} & 0.72\std{0.0}\gooddelta{-0.21} & 1.04\std{0.0} & 1.05\std{0.0}\baddelta{+0.01}\\
        \rowcolor{methodblue}\method & 0.26\std{0.03} & 0.16\std{0.01}\gooddelta{-0.10} & 0.22\std{0.01} & 0.19\std{0.01}\gooddelta{-0.03} & 0.59\std{0.0} & 0.59\std{0.0}\gooddelta{$\pm$0.0} \\
        \bottomrule
    \end{tabular}
    \begin{tabular}{l| cc | cc | cc}
        \toprule
        \multirow{2}*{method} & \multicolumn{2}{c}{Static DAVIS2017 $\downarrow$}& \multicolumn{2}{c}{Static YouTubeVOS $\downarrow$} & \multicolumn{2}{c}{CameraBench $\uparrow$} \\
        & T=2 & T=4 & T=2 & T=4 & T=2 & T=5\\
        \midrule
        baseline & 0.78\std{0.03} & 1.02\std{0.0}\baddelta{+0.24} & 0.91\std{0.0} & 0.92\std{0.0}\baddelta{+0.01} & 67.1\std{2.2} & 69.2\std{0.9}\gooddelta{+2.1}\\
        \rowcolor{methodblue}\method &0.87\std{0.07} & 0.87\std{0.05}\gooddelta{$\pm$0.0} & 0.70\std{0.01} & 0.71\std{0.02}\baddelta{+0.01} & 84.7\std{1.1} & 85.3\std{1.7}\gooddelta{+0.6} \\
        \bottomrule
    \end{tabular}
\end{table*}

\begin{table*}[ht]
    \caption{\textbf{Additional temporal-context results on SSv2-$110$k.} SDM improves monotonically as the number of input frames increases, including beyond the $T=5$ sequence length used during training. While the frame-pair baseline also improves with increasing temporal context, even with the maximum number of $T=7$ frames it performs worse than \method on a single pair of frames.
    Accuracy is reported in \%.}
    \label{tab:temporal_context_ssv2_appendix}
    \centering
    \small
    \setlength{\tabcolsep}{0.4em}
    \begin{tabular}{l| cccccc }
        \toprule
        \multirow{2}*{method} & \multicolumn{6}{c}{SSv2-$110$k}\\
        & T=2 & T=3 & T=4 & T=5 & T=6 & T=7\\
        \midrule
        baseline & 2.4\std{0.2} & 5.9\std{0.0}\gooddelta{+3.5} & 8.4\std{0.0}\gooddelta{+2.5} & 10.9\std{0.2}\gooddelta{+2.5} & 11.7\std{0.0}\gooddelta{+0.8} & 13.3\std{0.0}\gooddelta{+1.6}\\
        \rowcolor{methodblue}\method &16.7\std{0.7} & 18.6\std{0.8}\gooddelta{+1.9} & 20.4\std{1.1}\gooddelta{+1.8} & 21.6\std{1.2}\gooddelta{+1.2} & 22.5\std{1.7}\gooddelta{+0.9} & 23.1\std{1.3}\gooddelta{+0.6}\\
        \bottomrule
    \end{tabular}
\end{table*}

\subsection{Structured motion}\label{sec:additional_structured_results_appendix}
\Cref{tab:structured_motion_appendix} reports the full structured-motion results with standard deviations. The primary and residual tokens specialize according to the dominant motion source. The primary token performs best for camera motion in dynamic-camera videos, for object motion in approximately static-camera videos, and for SSv2-$110$k action prediction, while the residual token is stronger for Kubric object motion.
One exception is static DAVIS2017, where the residual token outperforms the primary token. This result is highly variable across seeds and therefore less reliable.
The performance of the upper-bound probes based on the concatenated $p$ and $r$ tokens remains close to the best performance achieved with a single token on most benchmarks. This indicates that most of the task-relevant information is encoded in a single token. A large gain appears on Kubric object motion, where predicting object translation in world coordinates can benefit from both the residual object-motion signal and the primary camera-motion component.

\begin{table*}[ht]
    \caption{\textbf{Additional structured-motion results.} 
        Full results with standard deviations showing that SDM learns specialized primary and residual motion tokens. 
        The primary token captures the dominant motion source, while the residual token captures remaining dynamics. 
        The only exception is static DAVIS2017, where the residual token outperforms the primary token.
        However, the high variability of this result makes it less reliable.
        We report MSE on standardized regression targets and accuracy in \% for classification tasks. 
        Among individual tokens, the best result is shown in \textbf{bold}. 
        Representations and targets that match are highlighted in \colorbox{matchcolour}{green}.
        The last row reports an upper-bound probe using the concatenated tokens $[p,r]$.
    }
    \label{tab:structured_motion_appendix}
    \centering
    \small
    \setlength{\tabcolsep}{0.4em}
    \begin{tabular}{l| ccccc c cc }
        \toprule
        \multirow{3}*{token} & \multicolumn{5}{c}{Regression (MSE $\downarrow$)} && \multicolumn{2}{c}{Classification (accuracy $\uparrow$)}\\
        \cmidrule{2-6} \cmidrule{8-9}
        & Kubric & Kubric & \multirow{2}*{DL3DV}& Static & Static && Camera & \multirow{2}*{SSv2-$110$k}\\
        & camera & object & & DAVIS2017 & YouTubeVOS && Bench & \\
        \midrule
        primary & \cellcolor{matchcolour}\textbf{0.16}\std{0.01} & 0.21\std{0.03} & \cellcolor{matchcolour}\textbf{0.59}\std{0.0} & \cellcolor{matchcolour}0.87\std{0.05} & \cellcolor{matchcolour}\textbf{0.71}\std{0.02} && \cellcolor{matchcolour}\textbf{85.3}\std{1.7} & \cellcolor{matchcolour}\textbf{23.1}\std{1.3}\\
        residual & 0.28\std{0.04} & \cellcolor{matchcolour}\textbf{0.19}\std{0.01} & 0.62\std{0.0} & \textbf{0.80}\std{0.15} & 0.79\std{0.01} && 81.1\std{1.9} & 12.0\std{0.9} \\
        \midrule\midrule
        concat. $[p,r]$ & 0.16\std{0.01} & 0.14\std{0.01} & 0.56\std{0.0} & 0.81\std{0.05} & 0.71\std{0.00} && 88.0\std{0.3} & 22.1\std{1.2}\\
        \bottomrule
    \end{tabular}
\end{table*}

\subsection{Ablation}\label{sec:additional_ablation_results_appendix}
\Cref{tab:ablation_appendix} provides the detailed component ablations summarized in \Cref{sec:ablation}.

\textbf{Weak supervision and sequential extraction lead to specialization.}
On Kubric, \method yields the expected specialization: $p$ is better for camera motion ($0.16$ vs.\ $0.28$), while $r$ is better for object motion ($0.19$ vs.\ $0.21$). 
Removing static-scene supervision weakens this structure, making the residual token nearly as predictive of camera motion as the primary token ($0.21$ vs.\ $0.18$).
Removing static-camera supervision has a stronger effect on object motion. $p$ becomes more predictive than $r$ ($0.14$ vs.\ $0.38$), indicating that the static-camera constraint is important for preventing object motion from being absorbed into the primary token. 
Jointly extracting $p$ and $r$ also weakens specialization, with $p$ becoming more predictive for object than camera motion.

\textbf{Intermediate input features help across benchmarks.}
Using all encoder layers instead of only final-layer features improves performance across all benchmarks, including Kubric camera error from $0.50$ to $0.16$, DL3DV error from $0.89$ to $0.59$, and SSv2-$110$k accuracy from $13.6\%$ to $23.1\%$.
Despite increasing the raw feature dimensionality by $12{\times}$, the fused prediction space keeps training efficient, increasing training time by only $1.2{\times}$.

\textbf{Unlabeled real data improves generalization.}
Adding unlabeled real videos improves over synthetic-only training on all evaluated real benchmarks: DL3DV error decreases from $0.91$ to $0.59$, while CameraBench and SSv2-$110$k accuracy increase from $76.6\%$ to $85.3\%$ and from $10.7\%$ to $23.1\%$, respectively.

\textbf{MSE improves Kubric object probing.}
Compared to $\mathcal{L}_1$, \emph{MSE} improves Kubric object-motion error from $0.30$ to $0.19$ and Kubric camera-motion error from $0.18$ to $0.16$. In contrast, $\mathcal{L}_1$ performs slightly better on DL3DV ($0.57$ vs.\ $0.59$), CameraBench ($87.4\%$ vs.\ $85.3\%$), and SSv2-$110$k ($24.4\%$ vs.\ $23.1\%$). We use \emph{MSE} because it significantly improves Kubric object-motion performance.

\textbf{Static-camera regularization balances object motion and action prediction.}
We vary $\lambda_{\mathrm{reg}}$, the weight of the static-camera regularization term. 
Smaller values yield stronger SSv2-$110$k performance, while larger values improve Kubric object-motion probing. 
We use $\lambda_{\mathrm{reg}}=0.5$ as a compromise, preserving strong Kubric object-motion performance ($0.19$) and SSv2-$110$k accuracy ($23.1\%$).

\begin{table*}[ht]
    \caption{\textbf{Detailed component ablations.}
        We ablate weak scene-level supervision, sequential primary/residual extraction, intermediate encoder features, unlabeled real-video training, the feature-prediction loss, and the regularization strength. We report MSE on standardized regression targets and accuracy in \% for classification tasks. 
        \colorbox{methodblue}{\strut Blue} indicates settings used by \method.
        Best performance for each metric is highlighted in \textbf{bold}.
    }
    \label{tab:ablation_appendix}
    \begin{subtable}[h]{\textwidth}
        \caption{\textbf{Weak supervision and sequential extraction improve token specialization.} Matched token--target pairs are highlighted in \colorbox{matchcolour}{green}.}
        \label{tab:structured_motion_ablation}
        \centering
        \footnotesize
        \setlength{\tabcolsep}{0.35em}
        \begin{tabular}{l| cc | cc | cc | cc}
            \toprule
            \multirow{3}*{token} & \multicolumn{2}{c|}{\cellcolor{methodblue}\method} & \multicolumn{2}{c|}{no static scene sup.} & \multicolumn{2}{c|}{no static camera sup.} & \multicolumn{2}{c}{joint motion extraction} \\
            \cmidrule{2-9}
             & Kubric & Kubric & Kubric & Kubric & Kubric & Kubric & Kubric & Kubric\\
            & camera $\downarrow$ & object $\downarrow$ & camera $\downarrow$ & object $\downarrow$ & camera $\downarrow$ & object $\downarrow$ & camera $\downarrow$ & object $\downarrow$\\
            \midrule
            primary & \cellcolor{matchcolour}\textbf{0.16} & 0.21 & \cellcolor{matchcolour}\textbf{0.18} & 0.22 & \cellcolor{matchcolour} \textbf{0.20} & \textbf{0.14} & \cellcolor{matchcolour} \textbf{0.19} & 0.16 \\
            residual & 0.28 & \cellcolor{matchcolour}\textbf{0.19} & 0.21 &\cellcolor{matchcolour}\textbf{0.20} & 0.44 &\cellcolor{matchcolour} 0.38 & 0.23 &\cellcolor{matchcolour} \textbf{0.15} \\
            \bottomrule
        \end{tabular}
    \end{subtable}
    \vspace{2mm}\\
    \begin{subtable}[ht]{0.49\textwidth}
        \caption{\textbf{Intermediate features improve performance.}}
        \label{tab:intermediate_feature_ablation}
        \centering
        \footnotesize
        \setlength{\tabcolsep}{0.35em}
        \begin{tabular}{l| ccccc}
            \toprule
            &\multicolumn{2}{c}{Kubric$\downarrow$}&\multirow{2}*{DL3DV$\downarrow$} & \multicolumn{1}{c}{Camera} &SSv2-\\
            &camera & object &&\multicolumn{1}{c}{Bench$\uparrow$} &  $110$k$\uparrow$\\
            \midrule
            only final feat. & 0.50 & 0.22 & 0.89 & 76.3 & 13.6\\
            \midrule
            \rowcolor{methodblue}multiple feat. & \textbf{0.16} & \textbf{0.19} & \textbf{0.59} & \textbf{85.3} & \textbf{23.1}\\
            \bottomrule
        \end{tabular}
    \end{subtable}
    \hfill
    \begin{subtable}[ht]{0.49\textwidth}
        \caption{\textbf{Unlabeled real data facilitates real-video generalization.}}
        \label{tab:real_data_ablation}
        \centering
        \footnotesize
        \setlength{\tabcolsep}{0.35em}
        \begin{tabular}{l| ccccc}
            \toprule
            &\multicolumn{2}{c}{Kubric$\downarrow$}&\multirow{2}*{DL3DV$\downarrow$} & \multicolumn{1}{c}{Camera} &SSv2-\\
            &camera & object &&\multicolumn{1}{c}{Bench$\uparrow$} &  $110$k$\uparrow$\\
            \midrule
            no real data & 0.17 & \textbf{0.11} & 0.91 & 76.6 & 10.7\\
            \midrule
            \rowcolor{methodblue}syn\&real data & \textbf{0.16} & 0.19 & \textbf{0.59} & \textbf{85.3} & \textbf{23.1}\\
            \bottomrule
        \end{tabular}
    \end{subtable}
    \vspace{2mm}\\
    \begin{subtable}[ht]{0.49\textwidth}
        \caption{\textbf{MSE loss improves Kubric object probing.}}
        \label{tab:loss_ablation}
        \centering
        \footnotesize
        \setlength{\tabcolsep}{0.35em}
        \begin{tabular}{l| ccccc}
            \toprule
            &\multicolumn{2}{c}{Kubric$\downarrow$}&\multirow{2}*{DL3DV$\downarrow$} & \multicolumn{1}{c}{Camera} &SSv2-\\
            &camera & object &&\multicolumn{1}{c}{Bench$\uparrow$} &  $110$k$\uparrow$\\
            \midrule
            $L_1$ & 0.18 & 0.30 & \textbf{0.57} & \textbf{87.4} & \textbf{24.4}\\
            \midrule
            \rowcolor{methodblue}MSE & \textbf{0.16} & \textbf{0.19} & 0.59 & 85.3 & 23.1\\
            \bottomrule
        \end{tabular}
    \end{subtable}
    \hfill
    \begin{subtable}[ht]{0.49\textwidth}
        \caption{\textbf{Medium static-camera regularization balances object motion and action prediction.}}
        \label{tab:reg_weight_ablation}
        \centering
        \footnotesize
        \setlength{\tabcolsep}{0.35em}
        \begin{tabular}{l| ccccc}
            \toprule
            &\multicolumn{2}{c}{Kubric$\downarrow$}&\multirow{2}*{DL3DV$\downarrow$} & \multicolumn{1}{c}{Camera} &SSv2-\\
            &camera & object &&\multicolumn{1}{c}{Bench$\uparrow$} &  $110$k$\uparrow$\\
            \midrule
            $\lambda_{\mathrm{reg}}=0.1$ & \textbf{0.16} & 0.41 & \textbf{0.56} & 86.3 & \textbf{25.1}\\
            $\lambda_{\mathrm{reg}}=0.25$ & 0.17 & 0.35 & \textbf{0.56} & \textbf{87.0} & 24.3\\
            $\lambda_{\mathrm{reg}}=1.0$ & 0.20 & \textbf{0.14} & 0.59 & 86.7 & 17.7\\
            \midrule
            \rowcolor{methodblue}$\lambda_{\mathrm{reg}}=0.5$ & \textbf{0.16} & 0.19 & 0.59 & 85.3 & 23.1\\
            \bottomrule
        \end{tabular}
    \end{subtable}\\
\end{table*}

\subsection{Additional visualization}
\begin{figure}[ht]
  \centering
  \includegraphics[width=\textwidth]{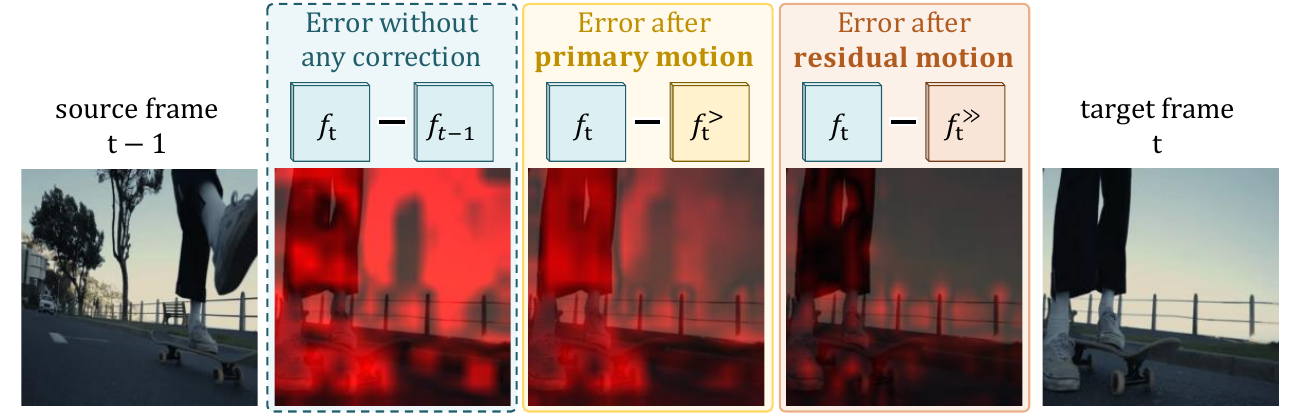}
  \caption{\textbf{Structured feature prediction.}
    We visualize feature-map error to the target after primary and residual compensation of a CameraBench sample. Errors are highlighted in red. 
    Compared to the error without any compensation, the \colorbox{prim}{\strut primary}-motion-compensated feature map reduces the error significantly for camera-motion-induced differences, particularly in the background.
    The \colorbox{res}{\strut residual} compensation further decreases the residual error and compensates for the error introduced by scene dynamics.
  }
  \label{fig:motion_compensated_fmaps_appendix}
\end{figure}

\Cref{fig:motion_compensated_fmaps_appendix} shows an additional visualization of the error between predicted and ground-truth feature maps.

\section{Limitations}\label{sec:limitations}
While a large portion of our training is fully self-supervised, the decomposition part is also anchored by weak scene-level supervision on synthetic data. Replacing this with fully self-supervised constraints would improve scalability. Finally, \benchmark relies on estimated camera motion for constructing static DAVIS2017 and YouTubeVOS, which can introduce filtering noise, and linear probes only measure linearly decodable information.

\section{Information about used assets}\label{sec:used_assets_appendix}
\Cref{tab:existing_assets} lists more detailed information about the pretrained models and datasets used.

\begin{table*}[ht]
    \caption{\textbf{Existing assets.}
    Summary of existing datasets and pretrained models used in this paper.}
    \label{tab:existing_assets}
    \centering
    \small
    \setlength{\tabcolsep}{0.4em}
    \begin{tabular}{l p{0.2\linewidth} p{0.1\linewidth} p{0.20\linewidth} l}
        \toprule
        Asset & Use in this paper & Version / subset & License / terms & Citation \\
        \midrule
        \subtitlerow{5}{Pretrained models}
        DINOv2-reg-B/14 & Main frozen image encoder & public checkpoint & \href{https://github.com/facebookresearch/dinov2/blob/7b187bd4df8efce2cbcbbb67bd01532c19bf4c9c/LICENSE}{Apache License 2.0} & \cite{darcet2024dino_registers} \\
        MAE-B/16 & Frozen encoder for backbone ablation & public checkpoint & \href{https://huggingface.co/facebook/vit-mae-base}{Apache License 2.0} & \cite{he2022mae} \\
        SigLIP-B/16 & Frozen encoder for backbone ablation & public checkpoint &\href{https://huggingface.co/google/siglip-base-patch16-224}{Apache License 2.0} & \cite{zhai2023siglip} \\
        DINOv3-B/16 & Frozen encoder for backbone ablation & public checkpoint & \href{https://github.com/facebookresearch/dinov3/blob/31703e4cbf1ccb7c4a72daa1350405f86754b6d1/LICENSE.md}{DINOv3 License} & \cite{simeoni2025dinov3} \\
        DeltaTok & Self-supervised baseline & released pretrained checkpoint & \href{https://github.com/amazon-far/deltatok?tab=Apache-2.0-1-ov-file}{Apache License 2.0} & \cite{kerssies2026frame} \\
        VGGT-1B & Supervised 3D representation baseline and camera-motion filtering for static-camera benchmarks & public checkpoint & \href{https://huggingface.co/facebook/VGGT-1B}{CC BY-NC 4.0} & \cite{wang2025vggt} \\
        Depth Anything 3 (GIANT-1.1) & Supervised 3D representation baseline & public checkpoint & \href{https://github.com/ByteDance-Seed/Depth-Anything-3/blob/41736238f5bced4debf3f2a12375d2466874866d/README.md}{CC BY-NC 4.0} & \cite{lin2025depthanythingv3} \\
        Pi3X & Supervised 3D representation baseline & public checkpoint & \href{https://github.com/yyfz/Pi3/blob/b412c3bd236dfd7686f1e4b48004d5087f2fa093/README.md}{CC BY-NC 4.0} & \cite{wang2025pi3} \\
        \midrule
        \subtitlerow{5}{Datasets}
        Kubric / MOVi-E & training and evaluation & Custom generated clips based on MOVi-E setup &  \href{https://github.com/google-research/kubric/blob/4d5a0d4ee80cac1c318f58bed83db284f6c70036/LICENSE}{Apache License 2.0}  & \cite{greff2021kubric} \\
        Something-Something v2 & training and evaluation & Filtered train/validation split & \href{https://www.qualcomm.com/content/dam/qcomm-martech/dm-assets/documents/jester_something_something_exercise_research_license_final_qti_28jul2022.pdf}{license} & \cite{goyal2017ssv2} \\
        DL3DV & training and evaluation & \texttt{1K}--\texttt{7K} subsets & \href{https://github.com/DL3DV-10K/Dataset/blob/cda4445d0d987758c8a66f2ae868ba3ece6d2682/License.md}{DL3DV-10K Terms of use} & \cite{ling2024dl3dv} \\
        CameraBench & evaluation & Selected camera-motion class groups & \href{https://huggingface.co/datasets/syCen/CameraBench}{CC-BY-4.0} & \cite{lin2025camerabench} \\
        DAVIS2017 & evaluation & Official train/test split after filtering & \href{https://github.com/davisvideochallenge/davis2017-evaluation/blob/ac7c43fca936f9722837b7fbd337d284ba37004b/LICENSE}{BSD 3-Clause License} & \cite{pont2017davis2017} \\
        YouTubeVOS & evaluation & Video-level pseudo split after filtering & \href{https://youtube-vos.org/dataset/term/}{CC BY 4.0 (annotations) / non-commercial research purpose only (data)} & \cite{xu2018youtubevos1,xu2018youtubevos2} \\
        \bottomrule
    \end{tabular}
\end{table*}

\section{Hardware resources}\label{sec:hw_resources_appendix}
The experiments were conducted on a local cluster using a single node with $4$ NVIDIA H100 GPUs and $128$ CPU cores.
One \method training run takes approximately $11$ hours on such a node and uses up to 75GB of RAM and 25GB of VRAM per GPU. The evaluation on \benchmark takes approximately $25$ minutes.

The reported main results use three seeds, for a total of approximately $11\mathrm{h}\times 4 \times 3=132$ GPU-hours, where $4$ is the number of GPUs per node. The $9$ ablation runs require approximately $396$ GPU-hours.